\documentclass[review,authoryear]{elsarticle}

\usepackage{lineno,hyperref}
\modulolinenumbers[1]
\usepackage{amsmath}
\usepackage{amssymb}
\usepackage{epsfig}
\usepackage{graphicx}
\usepackage[ruled,lined]{algorithm2e}
\usepackage{subfigure}
\usepackage{url}
\usepackage{float}
\usepackage[compact]{titlesec}
\usepackage{caption}
\usepackage{setspace}
\usepackage{balance}
\usepackage{xcolor}
\usepackage{caption}
\usepackage{floatpag}
\usepackage{ulem}
\usepackage{multirow}
\usepackage{soul,color}

\journal{Neurocomputing}

\begin{document}

\begin{frontmatter}

\title{Exploring the Representational Power \\of Graph Autoencoder}

\author{Maroun Haddad and Mohamed Bouguessa}

\address{Department of Computer Science\\ University of Quebec at Montreal\\Montreal, Quebec, Canada\\haddad.maroun@uqam.ca, bouguessa.mohamed@uqam.ca}

\begin{abstract}

While representation learning has yielded a great success on many graph learning tasks, there is little understanding behind the structures that are being captured by these embeddings. For example, we wonder if the topological features, such as the Triangle Count, the Degree of the node and other centrality measures are concretely encoded in the embeddings. Furthermore, we ask if the presence of these structures in the embeddings is necessary for a better performance on the downstream tasks, such as  clustering and classification. To address these questions, we conduct an extensive empirical study over three classes of unsupervised graph embedding models and seven different variants of Graph Autoencoders.Our results show that five topological features: the Degree, the Local Clustering Score, the Betweenness Centrality, the Eigenvector Centrality, and Triangle Count are concretely preserved in the first layer of the graph autoencoder that employs the SUM aggregation rule, under the condition that the model preserves the second-order proximity. We supplement further evidence for the presence of these features by revealing a hierarchy in the distribution of the topological features in the embeddings of the aforementioned model. We also show that a model with such properties can outperform other models on certain downstream tasks, especially when the preserved features are relevant to the task at hand. Finally, we evaluate the suitability of our findings through a test case study related to social influence prediction.

\end{abstract}

\begin{keyword}
Graph Neural Networks \sep Graph Embedding \sep Node Representation Learning \sep Graph Topological Features \sep 
Explainable Artificial Intelligence.
\end{keyword}

\end{frontmatter}

% \linenumbers

\section{Introduction}

\subsection{Context}
Graph embedding approaches have emerged in recent decades as a powerful set of tools for learning representations on graphs, as well as for improving the performance on downstream tasks such as node clustering \citep{Wang2019}, classification \citep{Wu2019} and link prediction \citep{Ma2019}. While these embedding techniques do bring a lot of advantages over classical methods, such as learning with handcrafted features or direct learning on graphs, they do face a set of challenges that affect their quality and performance. Among these challenges, the greatest contributor to a “good” graph embedding is a vector representation that preserves the structure of the graph \citep{Goyal2018}. However, while many studies cite “the preservation of the graph structure by the embedding” as a requirement \citep{Goyal2018,Cai2018,Hamilton2017}, few works have attempted to investigate concretely this assumption and study its effect on the downstream learning tasks. To this end, this paper aims at answering the following important questions: (1) Are relevant topological structures being captured in the graph embeddings? (2) If yes, which embedding models best preserve these structures? and (3) What is the effect of preserving these structures on the downstream learning tasks? We believe that answering these questions will help us better understand and explain the content of the graph embeddings and the source of their representational power.
\newpage
It is important to note that the preservation of certain structural characteristics such as the orders of proximity is trivial, since the models that generate these embeddings are generally optimized to preserve a certain order of proximity between the embedded components \citep{Goyal2018,Cai2018}. This means that two components having a high value for a certain order of proximity between them, will have similar vector representations. However, the preservation of other important topological features such as the Degree of the node, its Local Clustering Score, and other centrality measures is yet to be studied in depth.

To address the questions put forward by this paper, we perform a detailed empirical study. First, we position the problem of graph embedding in the context of \textit{Unsupervised Learning} on \textit{attributed} graphs. Accordingly, we study the embeddings generated by three different classes of unsupervised graph embedding models: (1) Matrix Factorization, (2) Random Walk techniques, and (3) Graph Autoencoders (GAE). We further focus on Graph Autoencoders and study the embeddings generated by seven different variations of GAE. 

Second, we investigate the preservation of the topological features in the embeddings and pinpoint the models that best preserve these structures. We hypothesize, that if a certain topological feature can be approximated using the embeddings, then, the embeddings do encode certain information about the approximated feature. To this end, we adopt two strategies: (1) We directly predict the topological features with Linear Regression, using the embeddings as attributes \citep{Rizi2017}; (2) We transform the problem of preservation of the topological features into a classification problem \citep{Bonner2019}. Furthermore, to better understand the content of the embeddings, we visualize them in 2D space. This visualization will serve as an additional proof for the success of some models over others in preserving the topological features.

Third, we study the effect of the preservation of the topological features on the downstream tasks such as node clustering and  node classification. We believe finding a positive correlation between the preservation of the topological features and a good performance on the downstream tasks will be a strong argument for the necessity of having these features encoded in the embeddings. To this end, we evaluate the performance when the embeddings are used on three separate tasks: (1) We cluster the embeddings according to the ground truth of each dataset and measure the homogeneity of each cluster, (2) We evaluate  the  suitability  of  the  embeddings  on the  task  of  node  clustering, and (3) We evaluate the effect of the embeddings on the task of  node  classification. Finally, we evaluate the suitability of our findings through a test case study related to social influence prediction. The positive results of the case study would be an argument for the importance of having a “Topological Features Preservation” strategy when building a GNN architecture.

\subsection{Motivations and Contributions}
While graph embedding as a whole has received a great deal of interest in recent years, few studies have focused on the representational power of the embeddings or the structures that are being encoded in the vectors. Two closely related works to this paper are \citep{Rizi2017} and, \citep{Bonner2019}. In \citep{Rizi2017}, the authors only focused on the Random-Walk based techniques, where they attempted to use the embeddings as attributes to reconstruct the centrality measures of the nodes using linear regression. While the authors did find that the Closeness Centrality was being approximated to an extent by the embeddings, they had poor results for the preservation of the other topological features such as the Degree, Betweenness Centrality, and Eigenvector Centrality.

The authors in \citep{Bonner2019} hypothesized that if a mapping can be found between the embedding space and the topological features, then, these features are approximately captured in the embedding space. To this end, they transformed the problem of the preservation of topological features into a classification problem, where the centrality measures were divided into six classes using histogram binning. Subsequently, multiple classification models such as SVM, Logistic Regression, and Multi-Layer Neural Networks were used to predict the topological class of each node using the embeddings as attributes. However, their results were inconclusive on the majority of the tested benchmarks, except for the Eigenvector Centrality on the ego-Facebook dataset. Furthermore, the paper \citep{Bonner2019} missed studying the embeddings of the Graph Neural Network (GNN) models under the assumption that the GNN models only cover supervised learning. This assumption is, however, not necessarily true. Graph Autoencoders \citep{Kipf2016} have been widely used as an unsupervised learning variation of GNN for generating embeddings that hold state-of-the-art results on many unsupervised graph learning tasks \citep{Wang2019,Schlichtkrull2018,Hasanzadeh2019}.

In the domain of Graph Neural Networks, the Graph Isomorphism Network (GIN) \citep{Xu2019} was one of the first articles to study the expressive power of the SUM aggregation rule. In their seminal work, the authors in \citep{Xu2019} studied the capacity of several variants of  GNN at distinguishing different graph structures. By representing the features of the node neighbors as multisets, the authors argued that a maximally powerful GNN would always aggregate different multisets into different representations. Subsequently, they proposed GIN, a simple GNN that is as powerful as the Weisfeiler-Lehman graph isomorphism test. GIN uses the SUM rule to aggregate the features of the node neighbors and an MLP to map the aggregated features into a representation. Unlike the MEAN and MAX aggregators, the SUM aggregator can represent an injective multiset function, making it maximally powerful at distinguishing different multisets. Similarly, unlike the 1-layer perceptron, an MLP is a universal approximator of multiset functions, allowing it to map different multisets to different representations. However, in our work, we go beyond the capacity of the SUM aggregator at distinguishing general graph structures and study its ability to capture specific graph properties. Furthermore, we investigate the graph structures captured by each layer of the Graph autoencoder, contributing to a better explainability and understanding of the model functionality.

Lastly, the authors in \citep{Wu2019} developed a degree-specific GNN model that explicitly preserves the degree of the nodes in the embedding using multi-task learning. They found that the presence of the degree in the embedding did improve the performance of the model. However, the datasets used in \citep{Wu2019} to evaluate the model are generally skewed in favor of having the degree in the embedding, either by having the degree of the node as the ground-truth to be predicted, or by having the ground-truth labels heavily correlated with the degree of the node. In our study, we show that the degree and other topological features are naturally preserved in the first layer of the GNN that utilizes the SUM aggregation rule, without the need for explicitly preserving them. On the other hand, our results also suggest that the effect of having the topological features preserved in the embeddings is task-dependent, and may not always lead to a better performance on the downstream tasks.

Motivated by the inconclusive results of the previous studies \citep{Rizi2017,Bonner2019} and their limited scope in covering the most recent graph embedding models such as the GNN, we perform, in this paper, an extensive empirical study that covers the Graph Autoencoder models. Our experimental results reveal that the topological features are concretely preserved in the first layer of the Graph Autoencoders that employs the SUM aggregation rule. We further accentuate these findings by revealing an organized hierarchical structure in the distribution of the node degrees in the embeddings of this particular model. The visualized embeddings, depicted in the Experiments section, corroborate our claim.

To the extent of our knowledge, we are the first paper to study the preservation of the topological features for the Graph Autoencoders, and the first study to show substantial and conclusive results in favor of one type of model, across several benchmarks.

\vspace{0.5cm}

 We summarize the significance of our work as follows:
\begin{itemize}

    \item We investigate the representational power of graph embeddings in an extensive set of four test bed experiments that cover three different types of embedding models and seven different variants of Graph Autoencoders over eleven datasets.
    \item We empirically demonstrate that the topological features are best preserved in the first layer of the Graph Autoencoder that employs  the SUM aggregation rule under the condition that the model preserves the second-order proximity.
    \item We experimentally show that the presence of the topological features in the embeddings can generally improve the performance on the downstream task when the preserved features are relevant to the task.
    \item We show significant performance gains in a social influence prediction case study that leverages the findings of our experiments and demonstrates the importance of having a “Topological Features Preservation” strategy when building a GNN architecture.

\end{itemize}
\section{Unsupervised Graph Embedding}
While many different types of embeddings exist, including node \citep{Song2018}, edge \citep{Li2019}, cluster and entire graph embedding \citep{Dong2020}, in this study we focus on node embedding, as it is among the most widely used. Subsequently, for the rest of the paper, node embedding and graph embedding are used interchangeably.  
In the following, we explore three mainstream models for generating graph embeddings. 

\subsection{Matrix Factorization}
Some of the earliest methods for generating vector representations on graphs relied on Matrix Factorization. Laplacian Eigenmaps (LE) \citep{Belkin2001} is one of the first algorithms to apply this technique. LE aims to project nodes having a high first-order proximity between them into similar vectors in the embedding space. To embed the graph in $d$ dimensions, LE builds a similarity matrix and calculates the Laplacian of the graph and then derives its $d$ eigenvectors corresponding to the $d$ smallest eigenvalues. The embedding becomes the matrix $Y= [y_1,y_2,...,y_d]$ with the embedding of a node $v_i=Y^T_i$(the $i^{th}$ line  of $Y$).

\subsection{Random Walk Techniques}
The authors of Deepwalk \citep{Perozzi2014} adopted the same word embedding architecture used in Word2Vec \citep{Mikolov2013} in order to generate vector representations for the nodes in a graph. In the data preparation part, a series of random walks are performed on the graph in order to generate sequences of nodes equivalent to the sentences in Word2Vec. Next, a sliding window is glided on each sequence in order to generate segments of consecutive nodes that will serve as the training data of the model.

In the training part, a SkipGram model is applied on the data generated in the first section. The model attempts to predict the neighbors of each middle node in the sliding window. DeepWalk aims to project nodes having the same neighborhood (high second-order of proximity) into similar vectors in the embedding space. 

While DeepWalk uniformly samples the nodes during the random walk, Node2Vec \citep{Grover2016} is another variation of the same architecture that introduces two parameters $p$ and $q$ that control the sampling during the random walk. The parameter $p$ controls the probability of going back to a node $v$ after visiting another node $t$. The parameter $q$ controls the probability of moving away from node $v$ after visiting it.

\subsection{Graph Autoencoders}
The multi-layer neural networks in the domain of deep learning on graphs are generally referred to in literature as GNN (Graph Neural Networks). The GNN adopts the concept of message passing over the graph. In this technique, each node collects the messages, that is, the attributes of its direct neighborhood, which are then aggregated to form the new message of the node. 

Given a graph $G=(V,E, X)$, where $V$ is the set of nodes, $E$ the set of edges and $X$ the set of node attributes. The GNN takes two matrices as entry, the adjacency matrix $A \in\mathbb{R}^{|V|\times|V|}$, and the attributes matrix $X\in\mathbb{R}^{|V|\times k}$. In the case of an non-attributed graph, one-hot-encoding is used for messages instead of $X$. The main component of these networks is the forward propagation rule denoted by $\mathit{FWD}$. This function dictates the way the messages are being passed and aggregated in the network. $\mathit{FWD}$ takes two arguments as input, the adjancency matrix $A$ and the previous layer's hidden state $H_{k-1}$:

\begin{equation}
H_k=\mathit{FWD}(H_{k-1},A)
\end{equation}

$\mathit{FWD}$ is usually a non-linear function such as $ReLU$ or  $Sigmoid$. For the first layer, we have $H_0=X$ with the output of each layer serving as input for the next layer.  Every layer enables the exploration of a new level of neighborhood since the messages are accumulated from layer to layer. A weight matrix  $W$  is added to each layer to be learned and standard backpropagation is used for updating the weights.

The message passing and aggregation operation can be accomplished by multiplying the adjacency matrix with the node attributes matrix: $A\times X$. In order to accumulate the messages from layer to layer, self-loops are added to each node $\hat{A} = A+I$.

The following forward propagation rule aggregates the messages via SUM, where $\sigma$ is the activation function:
\begin{equation} 
H_k=\sigma(\hat{A} H_{k-1} W_k)
\end{equation}

In order to aggregate the MEAN of the messages, a normalized adjacency matrix can be used ($\hat{A}$ multiplied with the inverse degree matrix $\hat{D}^{-1/2}$):
\begin{equation} 
H_k=\sigma(\hat{D}^{-1} \hat{A}H_{k-1} W_k)
\end{equation}
\newpage
Another rule that was developed in \citep{Kipf2017} for the Graph Convolutional Network (GCN), is the spectral rule; it utilizes a renormalized symmetric adjacency matrix:
\begin{equation} 
H_k=  \sigma(\hat{D}^{-\frac{1}{2}} \hat{A}\hat{D}^{-\frac{1}{2}} H_{k-1} W_k)
\end{equation}

After introducing the GCN model, the authors in \citep{Kipf2016} extended the architecture into a Graph Autoencoder (GAE).

The encoder section of the GAE is built from GNN layers. Here is an example of a two-layer GAE using the spectral rule:
\begin{equation}
\begin{split}
    H_1 &=ReLU(\hat{D}^{-\frac{1}{2}}\hat{A}\hat{D}^{-\frac{1}{2}}X W_1)\\
    H_2 &= ReLU(\hat{D}^{-\frac{1}{2}}\hat{A}\hat{D}^{-\frac{1}{2}}H_1W_2)
\end{split}
\end{equation}
The decoder simply reconstructs the adjacency matrix  ($\bar{A}$). In order to reconstruct the matrix, the output of the encoder (latent space $Z = H_2$) is multiplied with its transpose and then a $Sigmoid$ is applied element-wise on the product:
\begin{equation}
\bar{A}=Sigmoid(ZZ^T)
\end{equation}

\section{Exploring The Power of Graph Autoencoders}

This section aims to deeply investigate the representational power of graph autoencoders. To this end, an extensive experimental protocol\footnote{We bring to the attention of the reader that more experimental results can also be found at: \url{https://github.com/MH-0/RPGAE}} is conducted to study the characteristics being captured by the embeddings and their effect on downstream tasks such as node clustering and classification. In order to perform this, we need to specify the scope of our experiments and the required experimental
configurations.
\subsection{Scope of the Experiments}
In order to achieve the goals put forward by this paper, we have laid out the four experiments of this study over three main steps. In the first step, we build multiple variations for each class of embedding model. This is so, in case one of the models is successful, we could pinpoint which elements contributed to its success against the other variations. Next, we perform the first experiment for the purpose of identifying the models that are successful in preserving the topological features, and highlight the reasons for their success. This experiment is further supplemented by a visualization section that serves as additional evidence for the success of some models in preserving the topological features over others. In the final step, we perform three experiments that evaluate the embeddings over three separate tasks (cluster homogeneity, node clustering, and node classification). For every experiment, we compare the performance of the models that preserved the topological features versus the models that did not. This comparison helps in highlighting the importance of the presence of the topological features in the embeddings. We conclude the experiments by an overall analysis of the results and recommendations for embedding model choices. 

\subsection{Datasets and Compared Baselines}

\begin{table}[t]
\fontsize{8}{8}\selectfont
\centering
\begin{tabular}{lcccc}
\hline\hline
{\color[HTML]{000000} \textbf{Dataset}} & {\color[HTML]{000000} \textbf{Nodes}} & {\color[HTML]{000000} \textbf{Edges}} & {\color[HTML]{000000} \textbf{Classes}} & {\color[HTML]{000000} \textbf{Type}}  \\ \hline
{\color[HTML]{000000} Cora} & {\color[HTML]{000000} 2,708} & {\color[HTML]{000000} 5,429} & {\color[HTML]{000000} 7} & {\color[HTML]{000000} Citation}\\
{\color[HTML]{000000} Citeseer} & {\color[HTML]{000000} 3,327} & {\color[HTML]{000000} 4,732} & {\color[HTML]{000000} 6} & {\color[HTML]{000000} Citation} \\
{\color[HTML]{000000} email-Eu-core} & {\color[HTML]{000000} 1,005} & {\color[HTML]{000000} 25,571} & {\color[HTML]{000000} 42} & {\color[HTML]{000000} Email} \\
{\color[HTML]{000000} USA Air-Traffic} & {\color[HTML]{000000} 1,190} & {\color[HTML]{000000} 13,599} & {\color[HTML]{000000} 4} & {\color[HTML]{000000} Flight} \\
{\color[HTML]{000000} Europe Air-Traffic} & {\color[HTML]{000000} 399} & {\color[HTML]{000000} 5,995} & {\color[HTML]{000000} 4} & {\color[HTML]{000000} Flight} \\
Brazil Air-Traffic & 131 & 1,074 & 4 & Flight \\
{\color[HTML]{000000} fly-drosophila-medulla-1} & {\color[HTML]{000000} 1,781} & {\color[HTML]{000000} 9,016} & {\color[HTML]{000000} NA} & {\color[HTML]{000000} Biological} \\
{\color[HTML]{000000} ego-Facebook} & {\color[HTML]{000000} 4,039} & {\color[HTML]{000000} 88,234} & {\color[HTML]{000000} NA} & {\color[HTML]{000000} Social} \\
{\color[HTML]{000000} soc-sign-bitcoin-alpha} & {\color[HTML]{000000} 3,783} & {\color[HTML]{000000} 24,186} & {\color[HTML]{000000} NA} & {\color[HTML]{000000} Blockchain} \\
soc-sign-bitcoin-otc & 5,881 & 35,592 & NA & Blockchain \\
{\color[HTML]{000000} ca-GrQc} & {\color[HTML]{000000} 5,242} & {\color[HTML]{000000} 14,496} & {\color[HTML]{000000} NA} & {\color[HTML]{000000} Collaboration} \\ 
\hline\hline
\end{tabular}
\caption{Graph Datasets.}
\label{tab:graphdataset}
\end{table}
\begin{figure*}
\centerline{\includegraphics[scale=0.65]{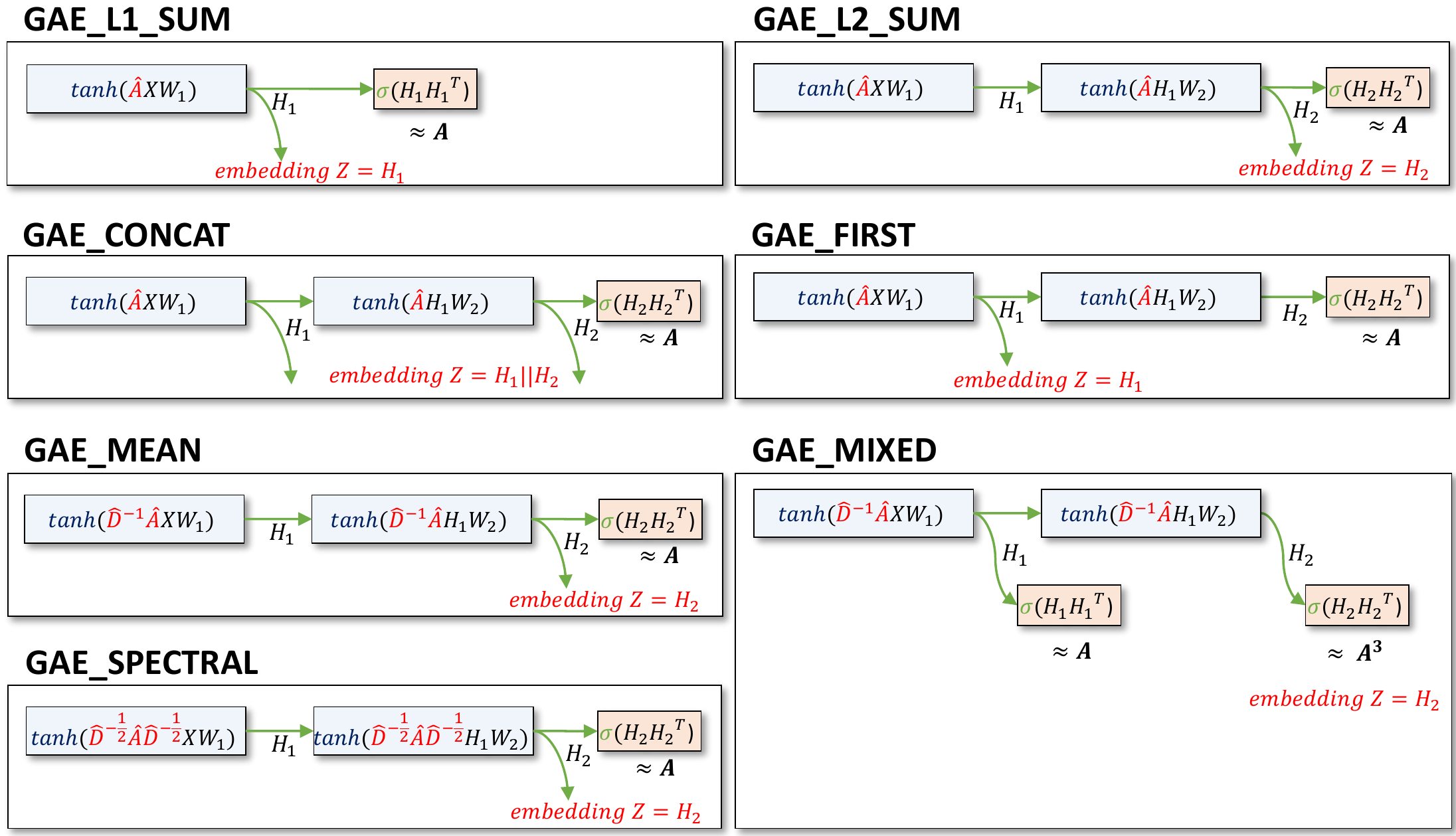}}
\caption{Seven variations of Graph Autoencoders.}
\label{fig:embeddingmodels}
\end{figure*}

We evaluate the embeddings generated by three types of architecture: (1) Matrix Factorization, (2) Random Walk and (3) Seven variations of Graph Autoencoders. For this purpose, we use eleven real datasets retrieved from the Stanford Large Network Dataset Collection \citep{snapnets}, Deep Graph Library \citep{Wang2019a} and \citep{Wu2019}. Table \ref{tab:graphdataset} summarizes the main characteristics of the collected datasets. Note that, Cora and Citeseer are attributed graph datasets. The dimensionality of Cora's attributes is 1,433 while the dimensionality of Citeseer's attributes is 3,703. The rest of the datasets are plain graphs where one-hot-encoding is used for attributes. 

\vspace{0.5cm}

\noindent In what follows, we briefly describe the compared baselines:

\textbf{Matrix Factorization}: We use the implementation for Laplacian Eigenmaps \citep{Belkin2001} included in Scikit-learn with the default parameters offered by the library.

\textbf{Random Walk}: We use two different variations of Node2Vec \citep{Grover2016} with the same setup presented in \citep{Bonner2019}. Node2Vec-Structural ($p=0.5, q = 2.0$) which explores global structures in the graph during the random walk and Node2Vec-Homophily ($p=1.0, q=0.5$) which explores local structures around the starting node. For both variations we use the parameters of the original paper (walk-length = 80 and number-walks = 10).
\vspace{0.3 cm}

\textbf{Graph Autoencoders}: We study seven different variations on the original architecture (see Figure \ref{fig:embeddingmodels}). We explore multiple aggregation rules: SUM \citep{Xu2019}, MEAN \citep{Hamilton2017a}, SPECTRAL \citep{Kipf2017}). We also look into different ways for selecting the embedding: Last layer output \citep{Kipf2016}, First layer output, or Concatenation of all layers outputs \citep{Xu2019}), as well as, the effect of reconstructing multiple orders of proximity. We adopt a two-layer architecture for all variations except when we mention otherwise.

\vspace{0.5cm}
\noindent\textbf{GAE\_L1\_SUM} is a graph autoencoder with a one-layer encoder using the SUM rule, with $Z$ as the embedding:
\begin{equation}
    Z = \tanh(\hat{A}XW_1)
\end{equation}
\textbf{GAE\_L2\_SUM} is a graph autoencoder with a two-layer encoder using the SUM rule:
\begin{equation}
\begin{split}
    H_1 &= \tanh(\hat{A}XW_1) \\
    H_2 &= \tanh(\hat{A}H_1W_2) \\
    Z &= H_2
\end{split}
\end{equation}
\newpage
\noindent\textbf{GAE\_CONCAT} is a graph autoencoder using the SUM rule. However, the embedding is formed by concatenating the output from all the layers of the encoder, with $H_1$ and $H_2$ the output of the first and second layer respectively:
\begin{equation}
\begin{split}
        H_1 &= \tanh(\hat{A}XW_1) \\ 
        H_2 &= \tanh(\hat{A}H_1W_2) \\ 
        Z &= H_1\mathbin\Vert H_2 
\end{split}
\end{equation}

\noindent\textbf{GAE\_FIRST} is a graph autoencoder using the SUM rule. However, the embedding is the output of the first layer of the encoder:
\begin{equation}
\begin{split}
      H_1 &= \tanh(\hat{A}XW_1) \\ 
      H_2 &= \tanh(\hat{A}H_1W_2) \\ 
      Z &= H_1
\end{split}
\end{equation}
\textbf{GAE\_MEAN} is a graph autoencoder using the MEAN rule:
\begin{equation}
\begin{split}
      H_1 &= \tanh(\hat{D}^{-1} \hat{A}XW_1) \\ 
      H_2 &= \tanh(\hat{D}^{-1} \hat{A}H_1W_2) \\ 
      Z &= H_2
\end{split}
\end{equation}
\textbf{GAE\_SPECTRAL} is a graph autoencoder that has a GCN as encoder and aggregates the messages using the SPECTRAL rule:
\begin{equation}
\begin{split}
H_1 &= \tanh(\hat{D}^{-1/2}\hat{A}\hat{D}^{-1/2}X W_1)  \\
H_2 &= \tanh(\hat{D}^{-1/2}\hat{A}\hat{D}^{-1/2}H_1W_2) \\
 Z &= H_2
\end{split}
\end{equation}
\textbf{GAE\_MIXED} is a graph autoencoder using the MEAN rule that reconstructs two orders of proximity. While the standard GAE reconstructs $A$. GAE\_MIXED reconstructs $A$ and $A^3$. We use the same encoder as GAE\_MEAN. However, the output of the first layer is used to reconstruct $A$, while the output of the second layer is used to reconstruct $A^3$. We have tested the reconstruction of multiple orders of proximity and selected $A$ and $A^3$ as they  gave the best results.

For the following models: GAE\_L1\_SUM, GAE\_L2\_SUM, GAE\_CONCAT, GAE\_FIRST,  GAE\_MEAN and GAE\_SPECTRAL the loss is the reconstruction error of the adjacency matrix $A$ calculated from the output of the last layer $H_l$:
\begin{equation}
  \mathcal{L} =  \frac{1}{|V|^2}\sum_{i=1}^{|V|}\sum_{j =1}^{|V|}(\sigma(H_l.H_l^{T}) - A)_{ij}^2  
\end{equation}

For GAE\_MIXED, the loss is the sum of both reconstruction errors of $A$ and $A^3$. The output of the first layer $H_1$ is used to reconstruct $A$ and the output of second layer $H_2$ is used to reconstruct $A^3$ with an attenuation factor alpha = 0.5:

\begin{equation}
  \mathcal{L} =  \frac{1}{|V|^2}\sum_{i=1}^{|V|}\sum_{j=1}^{|V|}(\sigma(H_1.H_1^{T}) - A)_{ij}^2
  + \alpha.\frac{1}{|V|^2}\sum_{i=1}^{|V|}\sum_{j=1}^{|V|}(\sigma(H_2.H_2^{T}) - A^3)_{ij}^2
\end{equation}

For the compared models, we use an embedding of size 64. For all graph autoencoders, we use a hidden layer of size 64 and an output layer of size 64 with hyperbolic tangent activation function and batch normalization on all layers without any dropout. We use an optimizer Adam with a learning rate of 0.01. We train for 250 epochs with a patience of 10. For the model GAE\_CONCAT we use layers of size 32 so the total would be 64. All experiments are run 10 times and we report the mean of the results. We regenerate the embeddings for each run.

\subsection{Topological Features Prediction}

We start our experiments by evaluating if the embeddings are capturing any of the graph topological features. We focus our study on the following five distinct features:

\begin{itemize}
    \item \textbf{Degree}  $DG(v)=K_v=deg(v)$ is the number of edges connected to a node $v$, counted twice for every self-loop.
    
    \item \textbf{Triangle Count} $TC(v) = \Delta_v$. The number of triangles of a node $v$. A tirangle or triplet $\Delta_v$ is a pair of neighbours of $v$ that are also connected among each other. 
    
    \item \textbf{Local clustering score} $LC(v) = \frac{2\Delta_v}{K_v\times(K_v-1)}$. It is the number of triangles or triplets $\Delta_v$ over the number of possible edges of $v$. The local clustering score measures the connectivity of the neighborhood of a node $v$. For a directed graph (such as email-Eu-core), the number of possible edges is $K_v\times(K_v-1)$ and $LC(v)=\frac{\Delta_v}{K_v\times(K_v-1)}$.
    \item \textbf{Eigenvector centrality}$EC(v) = \frac{1}{\lambda}\sum\limits_{n\in\mathcal{N}(v)}EC(n)$. Where $\lambda$ is a constant that is the largest eigenvalue. It measures the influence of a node calculated in reference to the influence of its neighborhood. It indicates that if the measure is high for $v$ then it is high for its neighbours.
    \item \textbf{Betweenness centrality} $BC(v) = \sum\limits_{s \neq t \neq v}\frac{\sigma_{st}(v)}{\sigma_{st}}$. Where $\sigma_{st}$ is the number of shortest routes between two nodes $s$ and $t$ and  $\sigma_{st}(v)$ is the number of shortest routes between $s$ and $t$ that pass through $v$. It measures the impact of node $v$ on the data transfer in the graph.
\end{itemize}

It is worth noting that, in  this paper,  we  do  not  aim at  defining new  features  to  reflect  the  topological  structure  of  a  graph.   This would be  far  beyond  the scope of our study.  Many topological features have already been proposed in the current literature.  In our experiments,  we utilize existing, well-known features  that  may  characterize  the  topological  structure  of  a  graph,  and  focus  on  evaluating if the embeddings  are  capturing any  of  the  graph  topological  features.  It is also important to note that we have used distinct features that investigate the topological structure of a graph from the perspective of the node, its neighbourhood, its role and importance in the graph.

To address the problem of the preservation of topological features, two approaches are possible: \citep{Rizi2017} and \citep{Bonner2019}. In the first approach \citep{Rizi2017}, the topological features are directly predicted with Linear Regression using the embeddings as attributes. A low prediction error indicates that the predicted feature is indeed encoded in the embeddings. In the second approach \citep{Bonner2019}, the problem is addressed as a classification problem. To this end, the range of values for each topological feature is divided, using histogram binning, into a set of intervals, where each interval of values corresponds to a topological class. This means that, in addition to its ground-truth class, each node will also have a Degree class, a Local Clustering Score class, an Eigenvector Centrality class, a Betweenness Centrality class, and a Triangle Count class. The topological class of the node indicates the interval into which its topological feature value falls.
While dividing the topological features into bins, we have tested over multiple number of bins and selected the ones that generated balanced classes.
In our study, in order to mitigate the limitations that any of the aforementioned approaches might have, and to further support our claims, we have adopted both approaches, that is, direct prediction and classification.

In order to reconstruct the topological features, we use Linear Regression (LN-R) evaluated with the Mean Square Error (MSE). A low value of MSE suggests a good result. For the classification of the nodes into their topological feature classes, we apply four models: Logistic Regression (LG-R), linear SVM (SVM-L), SVM with RBF Kernel (SVM-RBF)  and a MultiLayer Perceptron (MLP). Note that, for the MLP, we use two hidden layers with a ReLU activation function, an Adam optimizer and a 0.001 learning rate. We train for 200 epochs without early stopping. We used the implementation in Scikit-learn for all algorithms with a 5-fold cross validation. To evaluate LG-R, SVM-L, SVM-RBF and MLP, we report two evaluation metrics Macro-F1 and Micro-F1. The higher the values of these metrics the better the results. Given the large number of experiments for the task of topological features classification (5 topological features times 11 datasets) 
and in order to avoid encumbering the paper, we present a representative and diversified subset of the results. Recall that a complete list of all the results can be found at: \url{https://github.com/MH-0/RPGAE}

\begin{table}[!htbp]
\vspace*{-2.5cm}
\hspace*{-2.65cm}
\centering
\thisfloatpagestyle{empty}
\fontsize{6}{6}\selectfont
\setlength{\tabcolsep}{2pt}
\renewcommand{\arraystretch}{1}
\begin{tabular}{lccccccccc}
\hline\hline
\multicolumn{1}{c}{{\color[HTML]{000000} \textbf{}}} & \multicolumn{1}{c}{\textbf{LN-R}} & \multicolumn{2}{c}{{\color[HTML]{000000} \textbf{LG-R}}} & \multicolumn{2}{c}{{\color[HTML]{000000} \textbf{SVM-L}}} & \multicolumn{2}{c}{\textbf{SVM-RBF}} & \multicolumn{2}{c}{\textbf{MLP}} \\ \cline{2-10} 
\multicolumn{1}{c}{Models} & MSE & Macro-F1 & Micro-F1 & Macro-F1 & Micro-F1 & Macro-F1 & Micro-F1 & Macro-F1 & Micro-F1 \\ \hline
GAE\_FIRST & \multicolumn{1}{c|}{\textbf{0.107 \tiny ±0.023}} & \textbf{0.865 \tiny ±0.026} & \multicolumn{1}{c|}{\textbf{0.899 \tiny \tiny ±0.020}} & \textbf{0.789 \tiny \tiny ±0.046} & \multicolumn{1}{c|}{\textbf{0.837 \tiny \tiny ±0.037}} & \textbf{0.869 \tiny \tiny ±0.031} & \multicolumn{1}{c|}{\textbf{0.904 \tiny \tiny ±0.025}} & \textbf{0.768 \tiny \tiny ±0.03} & \textbf{0.850 \tiny ±0.018} \\
GAE\_CONCAT & \multicolumn{1}{c|}{0.152 \tiny ±0.035} & 0.818 \tiny ±0.035 & \multicolumn{1}{c|}{0.860 \tiny ±0.028} & 0.682 \tiny ±0.043 & \multicolumn{1}{c|}{0.737 \tiny ±0.041} & 0.765 \tiny ±0.038 & \multicolumn{1}{c|}{0.796 \tiny ±0.039} & 0.692 \tiny ±0.042 & 0.791 \tiny ±0.019 \\
GAE\_L1\_SUM & \multicolumn{1}{c|}{4.086 \tiny ±0.097} & 0.126 \tiny ±0.012 & \multicolumn{1}{c|}{0.400 \tiny ±0.005} & 0.062 \tiny ±0.006 & \multicolumn{1}{c|}{0.151 \tiny ±0.003} & 0.415 \tiny ±0.031 & \multicolumn{1}{c|}{0.446 \tiny ±0.033} & 0.155 \tiny ±0.015 & 0.415 \tiny ±0.008 \\
GAE\_L2\_SUM & \multicolumn{1}{c|}{3.478 \tiny ±0.134} & 0.232 \tiny ±0.013 & \multicolumn{1}{c|}{0.401 \tiny ±0.010} & 0.199 \tiny ±0.011 & \multicolumn{1}{c|}{0.306 \tiny ±0.018} & 0.238 \tiny ±0.011 & \multicolumn{1}{c|}{0.317 \tiny ±0.015} & 0.186 \tiny ±0.021 & 0.408 \tiny ±0.011 \\
GAE\_MEAN & \multicolumn{1}{c|}{4.160 \tiny ±0.055} & 0.155 \tiny ±0.008 & \multicolumn{1}{c|}{0.375 \tiny ±0.006} & 0.188 \tiny ±0.009 & \multicolumn{1}{c|}{0.262 \tiny ±0.014} & 0.156 \tiny ±0.008 & \multicolumn{1}{c|}{0.345 \tiny ±0.010} & 0.120 \tiny ±0.003 & 0.391 \tiny ±0.002 \\
GAE\_MIXED & \multicolumn{1}{c|}{4.137 \tiny ±0.053} & 0.151 \tiny ±0.009 & \multicolumn{1}{c|}{0.374 \tiny ±0.004} & 0.187 \tiny ±0.008 & \multicolumn{1}{c|}{0.253 \tiny ±0.019} & 0.158 \tiny ±0.008 & \multicolumn{1}{c|}{0.346 \tiny ±0.012} & 0.116 \tiny ±0.003 & 0.392 \tiny ±0.001 \\
GAE\_SPECTRAL & \multicolumn{1}{c|}{4.177 \tiny ±0.000} & 0.113 \tiny ±0.000 & \multicolumn{1}{c|}{0.394 \tiny ±0.000} & 0.138 \tiny ±0.014 & \multicolumn{1}{c|}{0.219 \tiny ±0.015} & 0.340 \tiny ±0.010 & \multicolumn{1}{c|}{0.415 \tiny ±0.013} & 0.115 \tiny ±0.004 & 0.394 \tiny ±0.002 \\
Matrix Factorization & \multicolumn{1}{c|}{4.177 \tiny ±0.000} & 0.113 \tiny ±0.000 & \multicolumn{1}{c|}{0.394 \tiny ±0.000} & 0.104 \tiny ±0.006 & \multicolumn{1}{c|}{0.170 \tiny ±0.003} & 0.134 \tiny ±0.006 & \multicolumn{1}{c|}{0.234 \tiny ±0.005} & 0.113 \tiny ±0.000 & 0.392 \tiny ±0.001 \\
Node2Vec-S & \multicolumn{1}{c|}{4.251 \tiny ±0.072} & 0.148 \tiny ±0.009 & \multicolumn{1}{c|}{0.358 \tiny ±0.005} & 0.175 \tiny ±0.006 & \multicolumn{1}{c|}{0.190 \tiny ±0.007} & 0.173 \tiny ±0.008 & \multicolumn{1}{c|}{0.287 \tiny ±0.013} & 0.130 \tiny ±0.007 & 0.381 \tiny ±0.006 \\
Node2Vec-H & \multicolumn{1}{c|}{4.165 \tiny ±0.084} & 0.154 \tiny ±0.006 & \multicolumn{1}{c|}{0.364 \tiny ±0.005} & 0.169 \tiny ±0.012 & \multicolumn{1}{c|}{0.181 \tiny ±0.013} & 0.173 \tiny ±0.009 & \multicolumn{1}{c|}{0.298 \tiny ±0.015} & 0.131 \tiny ±0.005 & 0.381 \tiny ±0.008 \\ \hline\hline
\end{tabular}
\vspace{-0.3cm}
\caption{Cora  - Degree.}
\label{tab:topofeatures1}
\vspace{0.5cm}

\hspace*{-2.65cm}
\centering
\begin{tabular}{lccccccccc}
\hline\hline
\multicolumn{1}{c}{{\color[HTML]{000000} \textbf{}}} & \multicolumn{1}{c}{\textbf{LN-R}} & \multicolumn{2}{c}{{\color[HTML]{000000} \textbf{LG-R}}} & \multicolumn{2}{c}{{\color[HTML]{000000} \textbf{SVM-L}}} & \multicolumn{2}{c}{\textbf{SVM-RBF}} & \multicolumn{2}{c}{\textbf{MLP}} \\ \cline{2-10} 
\multicolumn{1}{c}{Models}  & MSE & Macro-F1 & Micro-F1 & Macro-F1 & Micro-F1 & Macro-F1 & Micro-F1 & Macro-F1 & Micro-F1 \\ \hline
GAE\_FIRST & \multicolumn{1}{l|}{1.574 \tiny ±0.130} & 0.425 \tiny ±0.028 & \multicolumn{1}{l|}{0.551 \tiny ±0.022} & 0.292 \tiny ±0.043 & \multicolumn{1}{l|}{0.342 \tiny ±0.068} & \textbf{0.52 \tiny ±0.014} & \multicolumn{1}{l|}{\textbf{0.642 \tiny ±0.01}} & 0.386 \tiny ±0.019 & 0.557 \tiny ±0.017 \\
GAE\_CONCAT & \multicolumn{1}{l|}{\textbf{1.416 \tiny ±0.112}} & \textbf{0.435 \tiny ±0.027} & \multicolumn{1}{l|}{\textbf{0.562 \tiny ±0.018}} & \textbf{0.367 \tiny ±0.018} & \multicolumn{1}{l|}{\textbf{0.465 \tiny ±0.031}} & 0.502 \tiny ±0.010 & \multicolumn{1}{l|}{0.634 \tiny ±0.008} & \textbf{0.406 \tiny ±0.036} & \textbf{0.583 \tiny ±0.029} \\
GAE\_L1\_SUM & \multicolumn{1}{l|}{1.496 \tiny ±0.074} & 0.405 \tiny ±0.017 & \multicolumn{1}{l|}{0.551 \tiny ±0.014} & 0.292 \tiny ±0.028 & \multicolumn{1}{l|}{0.321 \tiny ±0.040} & 0.428 \tiny ±0.012 & \multicolumn{1}{l|}{0.606 \tiny ±0.006} & 0.331 \tiny ±0.030 & 0.538 \tiny ±0.018 \\
GAE\_L2\_SUM & \multicolumn{1}{l|}{1.694 \tiny ±0.109} & 0.346 \tiny ±0.020 & \multicolumn{1}{l|}{0.508 \tiny ±0.015} & 0.319 \tiny ±0.033 & \multicolumn{1}{l|}{0.453 \tiny ±0.044} & 0.322 \tiny ±0.022 & \multicolumn{1}{l|}{0.481 \tiny ±0.039} & 0.307 \tiny ±0.022 & 0.520 \tiny ±0.014 \\
GAE\_MEAN & \multicolumn{1}{l|}{1.816 \tiny ±0.090} & 0.317 \tiny ±0.017 & \multicolumn{1}{l|}{0.484 \tiny ±0.014} & 0.272 \tiny ±0.013 & \multicolumn{1}{l|}{0.392 \tiny ±0.026} & 0.266 \tiny ±0.021 & \multicolumn{1}{l|}{0.438 \tiny ±0.026} & 0.273 \tiny ±0.021 & 0.489 \tiny ±0.010 \\
GAE\_MIXED & \multicolumn{1}{l|}{1.651 \tiny ±0.074} & 0.343 \tiny ±0.014 & \multicolumn{1}{l|}{0.498 \tiny ±0.010} & 0.282 \tiny ±0.026 & \multicolumn{1}{l|}{0.395 \tiny ±0.027} & 0.325 \tiny ±0.022 & \multicolumn{1}{l|}{0.488 \tiny ±0.020} & 0.283 \tiny ±0.021 & 0.490 \tiny ±0.016 \\
GAE\_SPECTRAL & \multicolumn{1}{l|}{2.173 \tiny ±0.000} & 0.167 \tiny ±0.000 & \multicolumn{1}{l|}{0.500 \tiny ±0.000} & 0.187 \tiny ±0.010 & \multicolumn{1}{l|}{0.348 \tiny ±0.010} & 0.358 \tiny ±0.010 & \multicolumn{1}{l|}{0.508 \tiny ±0.008} & 0.167 \tiny ±0.000 & 0.500 \tiny ±0.000 \\
Matrix Factorization & \multicolumn{1}{l|}{2.173 \tiny ±0.000} & 0.167 \tiny ±0.000 & \multicolumn{1}{l|}{0.500 \tiny ±0.000} & 0.169 \tiny ±0.000 & \multicolumn{1}{l|}{0.294 \tiny ±0.000} & 0.182 \tiny ±0.002 & \multicolumn{1}{l|}{0.291 \tiny ±0.001} & 0.167 \tiny ±0.000 & 0.500 \tiny ±0.000 \\
Node2Vec-S & \multicolumn{1}{l|}{2.239 \tiny ±0.038} & 0.208 \tiny ±0.005 & \multicolumn{1}{l|}{0.459 \tiny ±0.006} & 0.236 \tiny ±0.012 & \multicolumn{1}{l|}{0.279 \tiny ±0.017} & 0.210 \tiny ±0.012 & \multicolumn{1}{l|}{0.449 \tiny ±0.020} & 0.187 \tiny ±0.009 & 0.470 \tiny ±0.010 \\
Node2Vec-H & \multicolumn{1}{l|}{2.235 \tiny ±0.040} & 0.206 \tiny ±0.008 & \multicolumn{1}{l|}{0.459 \tiny ±0.006} & 0.232 \tiny ±0.010 & \multicolumn{1}{l|}{0.280 \tiny ±0.008} & 0.210 \tiny ±0.008 & \multicolumn{1}{l|}{0.444 \tiny ±0.014} & 0.187 \tiny ±0.006 & 0.474 \tiny ±0.011 \\ \hline\hline
\end{tabular}
\vspace{-0.3cm}
\caption{fly-drosophila-medulla-1 - Local Clustering Score.}
\label{tab:topofeatures2}
\vspace{0.5cm}

\hspace*{-2.65cm}
\centering
\begin{tabular}{lccccccccc}
\hline\hline
\multicolumn{1}{c}{{\color[HTML]{000000} \textbf{}}} & \multicolumn{1}{c}{\textbf{LN-R}} & \multicolumn{2}{c}{{\color[HTML]{000000} \textbf{LG-R}}} & \multicolumn{2}{c}{{\color[HTML]{000000} \textbf{SVM-L}}} & \multicolumn{2}{c}{\textbf{SVM-RBF}} & \multicolumn{2}{c}{\textbf{MLP}} \\ \cline{2-10}
\multicolumn{1}{c}{Models}  & MSE & Macro-F1 & Micro-F1 & Macro-F1 & Micro-F1 & Macro-F1 & Micro-F1 & Macro-F1 & Micro-F1 \\ \hline
GAE\_FIRST & \multicolumn{1}{l|}{0.340 \tiny ±0.020} & 0.680 \tiny ±0.016 & \multicolumn{1}{l|}{0.683 \tiny ±0.016} & 0.644 \tiny ±0.020 & \multicolumn{1}{l|}{0.646 \tiny ±0.020} & 0.666 \tiny ±0.013 & \multicolumn{1}{l|}{0.668 \tiny ±0.013} & 0.354 \tiny ±0.021 & 0.439 \tiny ±0.021 \\
GAE\_CONCAT & \multicolumn{1}{l|}{\textbf{0.325 \tiny ±0.033}} & \textbf{0.703 \tiny ±0.022} & \multicolumn{1}{l|}{\textbf{0.705 \tiny ±0.021}} & \textbf{0.694 \tiny ±0.026} & \multicolumn{1}{l|}{\textbf{0.696 \tiny ±0.026}} & \textbf{0.692 \tiny ±0.013} & \multicolumn{1}{l|}{\textbf{0.692 \tiny ±0.013}} & \textbf{0.393 \tiny ±0.050} & \textbf{0.468 \tiny ±0.039} \\
GAE\_L1\_SUM & \multicolumn{1}{l|}{0.784 \tiny ±0.069} & 0.522 \tiny ±0.010 & \multicolumn{1}{l|}{0.528 \tiny ±0.010} & 0.468 \tiny ±0.017 & \multicolumn{1}{l|}{0.472 \tiny ±0.017} & 0.526 \tiny ±0.012 & \multicolumn{1}{l|}{0.540 \tiny ±0.011} & 0.307 \tiny ±0.024 & 0.392 \tiny ±0.023 \\
GAE\_L2\_SUM & \multicolumn{1}{l|}{0.784 \tiny ±0.056} & 0.540 \tiny ±0.015 & \multicolumn{1}{l|}{0.542 \tiny ±0.015} & 0.517 \tiny ±0.019 & \multicolumn{1}{l|}{0.520 \tiny ±0.018} & 0.498 \tiny ±0.017 & \multicolumn{1}{l|}{0.493 \tiny ±0.017} & 0.327 \tiny ±0.026 & 0.401 \tiny ±0.023 \\
GAE\_MEAN & \multicolumn{1}{l|}{0.601 \tiny ±0.043} & 0.573 \tiny ±0.010 & \multicolumn{1}{l|}{0.574 \tiny ±0.009} & 0.534 \tiny ±0.019 & \multicolumn{1}{l|}{0.536 \tiny ±0.019} & 0.507 \tiny ±0.017 & \multicolumn{1}{l|}{0.503 \tiny ±0.018} & 0.303 \tiny ±0.022 & 0.373 \tiny ±0.020 \\
GAE\_MIXED & \multicolumn{1}{l|}{0.735 \tiny ±0.046} & 0.521 \tiny ±0.013 & \multicolumn{1}{l|}{0.526 \tiny ±0.013} & 0.480 \tiny ±0.017 & \multicolumn{1}{l|}{0.485 \tiny ±0.017} & 0.491 \tiny ±0.013 & \multicolumn{1}{l|}{0.489 \tiny ±0.013} & 0.288 \tiny ±0.030 & 0.361 \tiny ±0.031 \\
GAE\_SPECTRAL & \multicolumn{1}{l|}{1.426 \tiny ±0.193} & 0.333 \tiny ±0.033 & \multicolumn{1}{l|}{0.415 \tiny ±0.019} & 0.123 \tiny ±0.032 & \multicolumn{1}{l|}{0.219 \tiny ±0.021} & 0.491 \tiny ±0.051 & \multicolumn{1}{l|}{0.509 \tiny ±0.040} & 0.113 \tiny ±0.040 & 0.226 \tiny ±0.042 \\
Matrix Factorization & \multicolumn{1}{l|}{4.372 \tiny ±0.000} & 0.191 \tiny ±0.000 & \multicolumn{1}{l|}{0.236 \tiny ±0.000} & 0.062 \tiny ±0.000 & \multicolumn{1}{l|}{0.174 \tiny ±0.000} & 0.380 \tiny ±0.000 & \multicolumn{1}{l|}{0.403 \tiny ±0.000} & 0.070 \tiny ±0.012 & 0.173 \tiny ±0.010 \\
Node2Vec-S & \multicolumn{1}{l|}{4.019 \tiny ±0.174} & 0.214 \tiny ±0.010 & \multicolumn{1}{l|}{0.237 \tiny ±0.010} & 0.145 \tiny ±0.008 & \multicolumn{1}{l|}{0.155 \tiny ±0.010} & 0.212 \tiny ±0.010 & \multicolumn{1}{l|}{0.266 \tiny ±0.009} & 0.130 \tiny ±0.019 & 0.200 \tiny ±0.014 \\
Node2Vec-H & \multicolumn{1}{l|}{4.268 \tiny ±0.212} & 0.199 \tiny ±0.013 & \multicolumn{1}{l|}{0.226 \tiny ±0.012} & 0.147 \tiny ±0.016 & \multicolumn{1}{l|}{0.156 \tiny ±0.018} & 0.213 \tiny ±0.014 & \multicolumn{1}{l|}{0.261 \tiny ±0.012} & 0.129 \tiny ±0.020 & 0.189 \tiny ±0.019 \\ \hline\hline
\end{tabular}
\vspace{-0.3cm}
\caption{email-Eu-core - Eigenvector Centrality.}
\label{tab:topofeatures3}
\vspace{0.5cm}

\hspace*{-2.65cm}
\centering
\begin{tabular}{lccccccccc}
\hline\hline
\multicolumn{1}{c}{{\color[HTML]{000000} \textbf{}}} & \multicolumn{1}{c}{\textbf{LN-R}} & \multicolumn{2}{c}{{\color[HTML]{000000} \textbf{LG-R}}} & \multicolumn{2}{c}{{\color[HTML]{000000} \textbf{SVM-L}}} & \multicolumn{2}{c}{\textbf{SVM-RBF}} & \multicolumn{2}{c}{\textbf{MLP}} \\ \cline{2-10}
\multicolumn{1}{c}{Models}  & MSE & Macro-F1 & Micro-F1 & Macro-F1 & Micro-F1 & Macro-F1 & Micro-F1 & Macro-F1 & Micro-F1 \\ \hline
GAE\_FIRST & \multicolumn{1}{l|}{0.595 \tiny ±0.042} & 0.572 \tiny ±0.02 & \multicolumn{1}{l|}{0.689 \tiny ±0.011} & 0.315 \tiny ±0.031 & \multicolumn{1}{l|}{0.382 \tiny ±0.036} & 0.524 \tiny ±0.016 & \multicolumn{1}{l|}{0.585 \tiny ±0.029} & \textbf{0.576 \tiny ±0.014} & 0.690 \tiny ±0.009 \\
GAE\_CONCAT & \multicolumn{1}{l|}{\textbf{0.529 \tiny ±0.037}} & \textbf{0.596 \tiny ±0.016} & \multicolumn{1}{l|}{\textbf{0.702 \tiny ±0.008}} & \textbf{0.360 \tiny ±0.053} & \multicolumn{1}{l|}{\textbf{0.454 \tiny ±0.070}} & \textbf{0.526 \tiny ±0.021} & \multicolumn{1}{l|}{\textbf{0.592 \tiny ±0.030}} & \textbf{0.576 \tiny ±0.025} & \textbf{0.696 \tiny ±0.012} \\
GAE\_L1\_SUM & \multicolumn{1}{l|}{1.975 \tiny ±0.086} & 0.264 \tiny ±0.018 & \multicolumn{1}{l|}{0.539 \tiny ±0.009} & 0.067 \tiny ±0.003 & \multicolumn{1}{l|}{0.148 \tiny ±0.006} & 0.130 \tiny ±0.024 & \multicolumn{1}{l|}{0.137 \tiny ±0.015} & 0.299 \tiny ±0.043 & 0.565 \tiny ±0.021 \\
GAE\_L2\_SUM & \multicolumn{1}{l|}{1.320 \tiny ±0.130} & 0.407 \tiny ±0.020 & \multicolumn{1}{l|}{0.576 \tiny ±0.014} & 0.317 \tiny ±0.022 & \multicolumn{1}{l|}{0.423 \tiny ±0.025} & 0.354 \tiny ±0.034 & \multicolumn{1}{l|}{0.429 \tiny ±0.041} & 0.397 \tiny ±0.022 & 0.583 \tiny ±0.012 \\
GAE\_MEAN & \multicolumn{1}{l|}{1.659 \tiny ±0.060} & 0.370 \tiny ±0.010 & \multicolumn{1}{l|}{0.536 \tiny ±0.007} & 0.278 \tiny ±0.023 & \multicolumn{1}{l|}{0.363 \tiny ±0.028} & 0.339 \tiny ±0.026 & \multicolumn{1}{l|}{0.501 \tiny ±0.018} & 0.333 \tiny ±0.016 & 0.544 \tiny ±0.017 \\
GAE\_MIXED & \multicolumn{1}{l|}{1.555 \tiny ±0.041} & 0.381 \tiny ±0.011 & \multicolumn{1}{l|}{0.545 \tiny ±0.007} & 0.293 \tiny ±0.024 & \multicolumn{1}{l|}{0.359 \tiny ±0.041} & 0.365 \tiny ±0.022 & \multicolumn{1}{l|}{0.507 \tiny ±0.023} & 0.353 \tiny ±0.013 & 0.555 \tiny ±0.009 \\
GAE\_SPECTRAL & \multicolumn{1}{l|}{2.333 \tiny ±0.000} & 0.167 \tiny ±0.000 & \multicolumn{1}{l|}{0.500 \tiny ±0.000} & 0.189 \tiny ±0.007 & \multicolumn{1}{l|}{0.315 \tiny ±0.017} & 0.355 \tiny ±0.006 & \multicolumn{1}{l|}{0.490 \tiny ±0.005} & 0.167 \tiny ±0.000 & 0.500 \tiny ±0.000 \\
Matrix Factorization & \multicolumn{1}{l|}{2.333 \tiny ±0.000} & 0.167 \tiny ±0.000 & \multicolumn{1}{l|}{0.500 \tiny ±0.000} & 0.074 \tiny ±0.000 & \multicolumn{1}{l|}{0.158 \tiny ±0.000} & 0.109 \tiny ±0.001 & \multicolumn{1}{l|}{0.192 \tiny ±0.001} & 0.166 \tiny ±0.001 & 0.492 \tiny ±0.008 \\
Node2Vec-S & \multicolumn{1}{l|}{2.341 \tiny ±0.023} & 0.188 \tiny ±0.004 & \multicolumn{1}{l|}{0.488 \tiny ±0.002} & 0.163 \tiny ±0.011 & \multicolumn{1}{l|}{0.192 \tiny ±0.010} & 0.136 \tiny ±0.006 & \multicolumn{1}{l|}{0.250 \tiny ±0.006} & 0.185 \tiny ±0.004 & 0.476 \tiny ±0.005 \\
Node2Vec-H & \multicolumn{1}{l|}{2.371 \tiny ±0.023} & 0.186 \tiny ±0.005 & \multicolumn{1}{l|}{0.485 \tiny ±0.003} & 0.181 \tiny ±0.010 & \multicolumn{1}{l|}{0.205 \tiny ±0.009} & 0.137 \tiny ±0.008 & \multicolumn{1}{l|}{0.259 \tiny ±0.009} & 0.186 \tiny ±0.007 & 0.476 \tiny ±0.006 \\ \hline\hline
\end{tabular}
\vspace{-0.3cm}
\caption{soc-sign-bitcoin-otc - Betweeness Centrality.}
\label{tab:topofeatures4}
\vspace{0.5cm}

\hspace*{-2.65cm}
\centering
\begin{tabular}{lccccccccc}
\hline\hline
\multicolumn{1}{c}{{\color[HTML]{000000} \textbf{}}} & \multicolumn{1}{c}{\textbf{LN-R}} & \multicolumn{2}{c}{{\color[HTML]{000000} \textbf{LG-R}}} & \multicolumn{2}{c}{{\color[HTML]{000000} \textbf{SVM-L}}} & \multicolumn{2}{c}{\textbf{SVM-RBF}} & \multicolumn{2}{c}{\textbf{MLP}} \\ \cline{2-10}
\multicolumn{1}{c}{Models}  & MSE & Macro-F1 & Micro-F1 & Macro-F1 & Micro-F1 & Macro-F1 & Micro-F1 & Macro-F1 & Micro-F1 \\ \hline
GAE\_FIRST & \multicolumn{1}{l|}{\textbf{0.082 \tiny ±0.014}} & \textbf{0.917 \tiny ±0.015} & \multicolumn{1}{l|}{\textbf{0.918 \tiny ±0.014}} & \textbf{0.910 \tiny ±0.017} & \multicolumn{1}{l|}{\textbf{0.911 \tiny ±0.017}} & \textbf{0.892 \tiny ±0.012} & \multicolumn{1}{l|}{\textbf{0.893 \tiny ±0.011}} & 0.534 \tiny ±0.113 & 0.619 \tiny ±0.097 \\
GAE\_CONCAT & \multicolumn{1}{l|}{0.111 \tiny ±0.015} & 0.888 \tiny ±0.015 & \multicolumn{1}{l|}{0.889 \tiny ±0.015} & 0.891 \tiny ±0.026 & \multicolumn{1}{l|}{0.892 \tiny ±0.025} & 0.832 \tiny ±0.028 & \multicolumn{1}{l|}{0.833 \tiny ±0.027} & \textbf{0.577 \tiny ±0.082} & \textbf{0.651 \tiny ±0.065} \\
GAE\_L1\_SUM & \multicolumn{1}{l|}{0.191 \tiny ±0.029} & 0.809 \tiny ±0.022 & \multicolumn{1}{l|}{0.814 \tiny ±0.022} & 0.804 \tiny ±0.026 & \multicolumn{1}{l|}{0.808 \tiny ±0.025} & 0.798 \tiny ±0.026 & \multicolumn{1}{l|}{0.800 \tiny ±0.025} & 0.508 \tiny ±0.085 & 0.586 \tiny ±0.063 \\
GAE\_L2\_SUM & \multicolumn{1}{l|}{0.237 \tiny ±0.045} & 0.768 \tiny ±0.037 & \multicolumn{1}{l|}{0.773 \tiny ±0.037} & 0.751 \tiny ±0.033 & \multicolumn{1}{l|}{0.758 \tiny ±0.032} & 0.693 \tiny ±0.032 & \multicolumn{1}{l|}{0.707 \tiny ±0.029} & 0.493 \tiny ±0.064 & 0.563 \tiny ±0.060 \\
GAE\_MEAN & \multicolumn{1}{l|}{0.258 \tiny ±0.031} & 0.752 \tiny ±0.024 & \multicolumn{1}{l|}{0.763 \tiny ±0.024} & 0.749 \tiny ±0.029 & \multicolumn{1}{l|}{0.756 \tiny ±0.029} & 0.739 \tiny ±0.023 & \multicolumn{1}{l|}{0.746 \tiny ±0.021} & 0.489 \tiny ±0.075 & 0.545 \tiny ±0.069 \\
GAE\_MIXED & \multicolumn{1}{l|}{0.323 \tiny ±0.025} & 0.712 \tiny ±0.028 & \multicolumn{1}{l|}{0.723 \tiny ±0.024} & 0.702 \tiny ±0.021 & \multicolumn{1}{l|}{0.712 \tiny ±0.019} & 0.770 \tiny ±0.023 & \multicolumn{1}{l|}{0.778 \tiny ±0.021} & 0.492 \tiny ±0.078 & 0.559 \tiny ±0.065 \\
GAE\_SPECTRAL & \multicolumn{1}{l|}{0.401 \tiny ±0.019} & 0.534 \tiny ±0.004 & \multicolumn{1}{l|}{0.655 \tiny ±0.004} & 0.240 \tiny ±0.006 & \multicolumn{1}{l|}{0.400 \tiny ±0.007} & 0.780 \tiny ±0.021 & \multicolumn{1}{l|}{0.790 \tiny ±0.019} & 0.207 \tiny ±0.047 & 0.365 \tiny ±0.036 \\
Matrix Factorization & \multicolumn{1}{l|}{1.491 \tiny ±0.071} & 0.303 \tiny ±0.024 & \multicolumn{1}{l|}{0.393 \tiny ±0.018} & 0.168 \tiny ±0.000 & \multicolumn{1}{l|}{0.336 \tiny ±0.000} & 0.818 \tiny ±0.000 & \multicolumn{1}{l|}{0.825 \tiny ±0.000} & 0.207 \tiny ±0.029 & 0.346 \tiny ±0.023 \\
Node2Vec-S & \multicolumn{1}{l|}{1.390 \tiny ±0.104} & 0.306 \tiny ±0.055 & \multicolumn{1}{l|}{0.332 \tiny ±0.052} & 0.283 \tiny ±0.047 & \multicolumn{1}{l|}{0.317 \tiny ±0.050} & 0.308 \tiny ±0.035 & \multicolumn{1}{l|}{0.345 \tiny ±0.031} & 0.252 \tiny ±0.045 & 0.336 \tiny ±0.048 \\
Node2Vec-H & \multicolumn{1}{l|}{1.391 \tiny ±0.158} & 0.306 \tiny ±0.050 & \multicolumn{1}{l|}{0.326 \tiny ±0.050} & 0.278 \tiny ±0.047 & \multicolumn{1}{l|}{0.317 \tiny ±0.047} & 0.255 \tiny ±0.043 & \multicolumn{1}{l|}{0.290 \tiny ±0.044} & 0.237 \tiny ±0.035 & 0.330 \tiny ±0.030 \\ \hline \hline
\end{tabular}
\vspace{-0.3cm}
\caption{Brazil Air-Traffic - Triangle Count.}
\label{tab:topofeatures5}
\end{table}
\newpage

Tables \ref{tab:topofeatures1}, \ref{tab:topofeatures2}, \ref{tab:topofeatures3},  \ref{tab:topofeatures4} and \ref{tab:topofeatures5} list the results for a combination of topological features and datasets. The first column of each table holds the results of the linear regression while the rest of the columns display the results of the classification. We found that GAE\_FIRST and GAE\_CONCAT outperform the other models by a large margin for both approaches (i.e., regression and classification), notably, over 60\%  for the prediction of the node Degree on the Cora dataset (Table \ref{tab:topofeatures1}). Based on the results of our extensive empirical study, we have observed that the success of these two models in preserving the features, especially the Degree of the nodes, can be attributed to their use of the SUM aggregation rule and the inclusion of the first layer output in the embedding.

In order to sustain the aforementioned observations, we provide in the following a principled analysis regarding the preservation of the degree in the first layer of the GNN when using the SUM rule. In our analysis, we consider two cases: (1) Non-attributed graphs, and (2) Attributed graphs.

Let $G$ be a graph with a set of nodes $V$. GAE is a graph autoencoder applied to $G$, with a GNN encoder that uses the SUM forward propagation rule  ($\mathit{FWD_{SUM}}$), such as:
\begin{equation} 
H_k=\mathit{FWD_{SUM}}(H_{k-1},\hat{A}) = \sigma(\hat{A} H_{k-1} W_k) 
\end{equation}

\vspace{-0.5cm}
\subsubsection*{\textbf{(1) Non-attributed graphs:}}

For non-attributed graphs, we use one-hot-encoding as input for the first layer, such as $H_0=I$, where $I$ is an identity matrix. Therefore, $H_1=\sigma(\hat{A} H_0 W_1) = \sigma(\hat{A} I W_1)$. Since $\hat{A}I=\hat{A}$, we have $H_1=\sigma(\hat{A}W_1)$.
This implies that the hidden state of a node $v: h_{1(v)}$ in layer 1 is formed from an aggregation over a set of weights from that layer. Furthermore, there is a one-to-one correspondence between the indices of the nodes in $A$ and the indices of the weight vectors in $W_1$ . Therefore, ignoring $\sigma$, we have:
\begin{equation}
\label{eq:sumweightindices}
 h_{1(v)} = \sum_{n \in \mathcal{N}_i(v)}w_{n}  = \sum_{\bar{w} \in \bar{W}_{\mathcal{N}_i(v)}} \bar{w}
\end{equation}
Where $\mathcal{N}_i(v)$ is the set of indices of the neighbors of $v$. $w_{n}$ is a weight vector in $W_1$ that corresponds to the $n^{th}$ neighbor of $v$. We further define $\bar{W}_{\mathcal{N}_i(v)}$ as the set of weights in $W_1$ whose indices correspond to the indices in $\mathcal{N}_i(v)$.

Let $ \bar{V} \subseteq V$ be a subset of nodes of $G$, where the nodes of $\bar{V}$ have high second-order proximity among each other. For example, all the nodes in $\bar{V}$ have the same label. We define $\mathcal{N}_i(\bar{V})$ as the set of all neighbor indices for the nodes in $\bar{V}$. $\bar{W}_{\mathcal{N}_i(\bar{V})}$ is the set of weights whose indices correspond to $\mathcal{N}_i(\bar{V})$. If GAE preserves the second-order proximity between the nodes of $\bar{V}$ in the embeddings, then the distance between the embedding of any pair of nodes in $\bar{V}$ is small, such as: $\forall (v_i,v_j)\in \bar{V}$ we have $dist(h_{1(v_i)},h_{1(v_j)})<\epsilon$, where $\epsilon$ is a small positive value close to zero.
Following Equation (\ref{eq:sumweightindices}), we have:
\begin{equation}
dist\bigg(\sum \bar{W}_{\mathcal{N}_i(v_i)},\sum \bar{W}_{\mathcal{N}_i(v_j)}\bigg)<\epsilon
\end{equation}
Since the nodes in $\bar{V}$ have a high second-order proximity among each other, then  $\forall (v_i,v_j)\in \bar{V}$ we have $\mathcal{N}_i(v_i) \simeq \mathcal{N}_i(v_j) \Rightarrow \bar{W}_{\mathcal{N}_i(v_i)} \simeq \bar{W}_{\mathcal{N}_i(v_j)}$. Since this is true  $\forall (v_i,v_j)\in \bar{V}$, then as $\epsilon \to 0 $ we observe that the weights in $\bar{W}_{\mathcal{N}_i(\bar{V})}$ converge closer to each other, such as: $\exists \bar{w_c}$, $\forall \bar{w} \in \bar{W}_{\mathcal{N}_i(\bar{V})}$ we have $dist(\bar{w},\bar{w_c})<\delta $. For $\delta >0$ and small. Then, $\forall v \in \bar{V}$ we have:
\begin{equation}
 h_{1(v)} = \sum_{n \in \mathcal{N}_i(v)}w_{n}  \approx |\mathcal{N}_i(v)| . \bar{w_c}
\end{equation}
This effect is however lost in the subsequent layers of the encoder, since $\hat{A} H_{k} \neq \hat{A}$ for $k>0$. That is why the vanilla GAE\_L2\_SUM that uses the output of the last layer in the encoder as embedding does not preserve the topological features. On the other hand, when we use the output of the first layer as the embedding or when we concatenate it with the output of the other layers, we preserve the topological features.

\subsubsection*{\textbf{(2) Attributed graphs:}}
For attributed graphs, we consider sparse Bag-of-words attributes, such as the ones used in our experiments with Cora and Citeseer datasets. In this case, we have a binary sparse attribute matrix $X\in\{0,1\}$. We observe that $\hat{A}X\approx \hat{A}$. Therefore, in case of a GAE with a SUM aggregation rule, we have $H_1=\sigma(\hat{A} H_0 W_1)=\sigma(\hat{A} X W_1) \approx \sigma(\hat{A}W_1)$. This leads to the same effect of aggregating over a converging set of weights, as described in the case of non-attributed graphs.

\vspace{0.3cm}

In both cases, of attributed and non-Attributed graphs, the condition of the preservation of the second-order proximity is crucial for the preservation of the topological features in the embeddings. We observe this fact in our experiments with the under-performance of GAE\_L1\_SUM (Autoencoder with one layer). Since it uses the SUM rule and the output of the first layer as embedding, it should have preserved the topological features. However, one layer is not sufficient for the GAE to converge on a dataset of the size of either Cora or Citeseer. This means that the weights associated to the neighbors of two nodes that have the same label, might not be similar. Aggregating these weights will not lead to similar embeddings for the two nodes even if they have the same number of neighbors. On the other hand, for relatively small networks such as USA Air-Traffic, or the ego-Facebook, one layer networks are sufficient for convergence and the topological features are preserved in GAE\_L1\_SUM. This is why the condition for the model convergence (i.e., the preservation of the second-order proximity) is necessary for the preservation of the topological features in the first layer.

\begin{figure*}[!t]
\centerline{\includegraphics[scale=0.40]{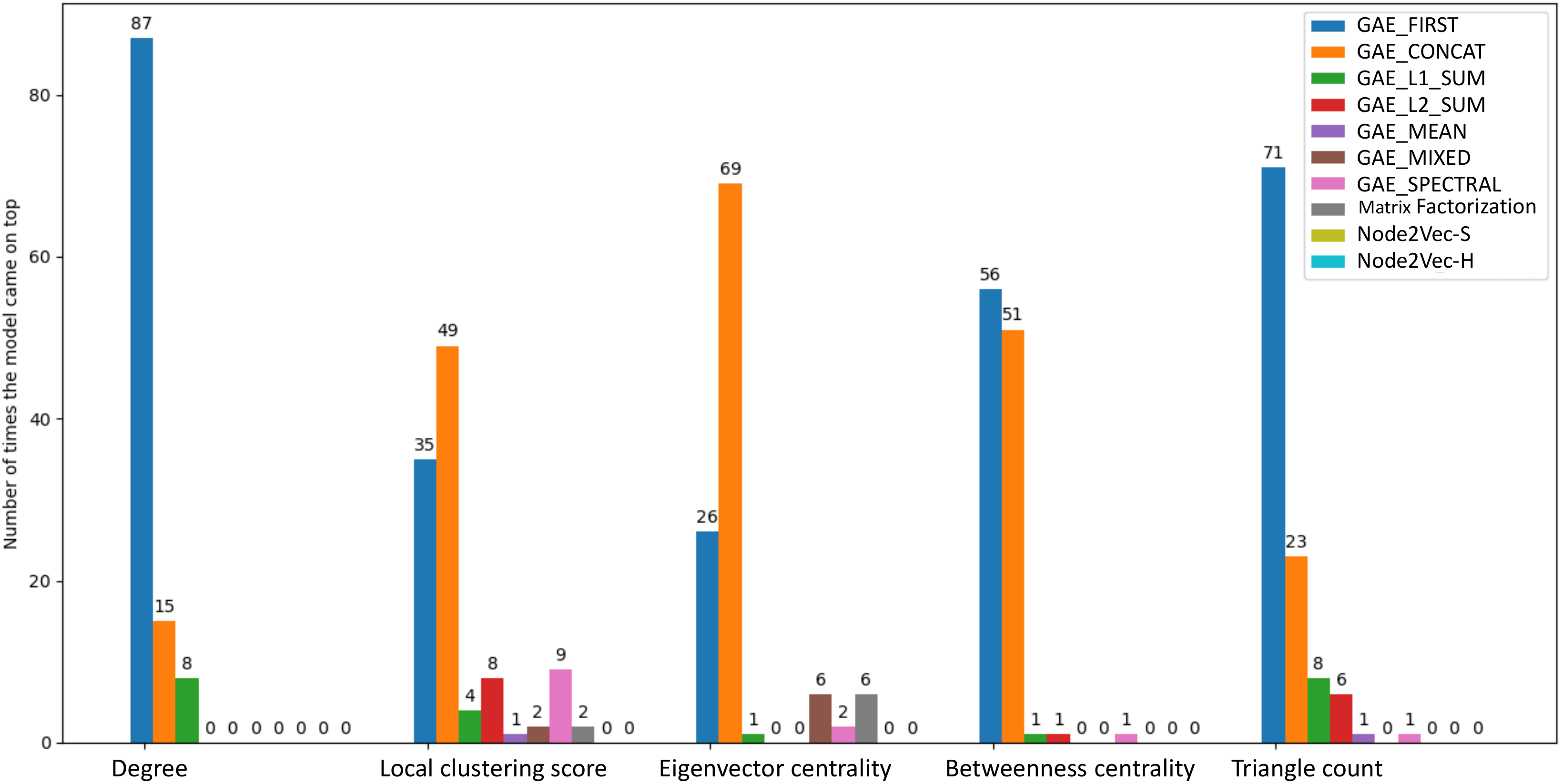}}
\caption{The number of times each model outperformed the others on the ten evaluation metrics over the eleven datasets per feature.}
\label{fig:bestmodels}
\end{figure*}

Furthermore, in order to have a better understanding of the performance of the models across the large number of experiments conducted in this work, we compile the results in a figure that illustrates the overall trends in the scores for the task of preservation of the topological features. Figure \ref{fig:bestmodels} shows the number of times each model outperformed the others for the ten evaluation metrics over the eleven datasets (that is, 11 x 10 = 110 evaluation scores per feature). The evaluation metrics are (MSE and MAE) for: Linear Regression. (F1-Macro and F1-Micro) for each of the four models: Logistic Regression (LG-R), Linear SVM (SVM-L), SVM with RBF Kernel (SVM-RBF), and Multilayer Perceptron (MLP).

Figure \ref{fig:bestmodels} clearly shows that GAE\_FIRST (dark blue) and GAE\_CONCAT (orange) outperform all other models by a large margin. In particular, it shows that GAE\_FIRST is favorable for capturing the Degree and the Triangle Count, while GAE\_CONCAT better captures the Local clustering score and the Eigenvector Centrality. As for the Betweeness Centrality, the results are shared between both GAE\_FIRST and GAE\_CONCAT. We believe that this is due to the nature of each feature. The Degree and Trianle Count of the node are related to its one-hop neighborhood. For this reason, the Degree and Triangle Count are best preserved in the first hidden layer of the autoencoder, since this layer aggregates the one-hop neighborhood features of the node. However, the Eigenvector Centrality and the Local Clustering Score  are two measures that depend on the two-hop neighborhood, that is, the importance and structure of the neighborhood of the node. The Eigenvector Centrality for a certain node is high, if it is also high for its neighbors. As for the Local Clustering Score, it measures the connectivity of the neighborhood. This is why these features are best preserved when we also include the output of  the second hidden layer in the embedding, since this layer captures information about the two-hop neighborhood of the node.

\begin{figure*}[!htbp]

\centerline{\includegraphics[scale=0.70]{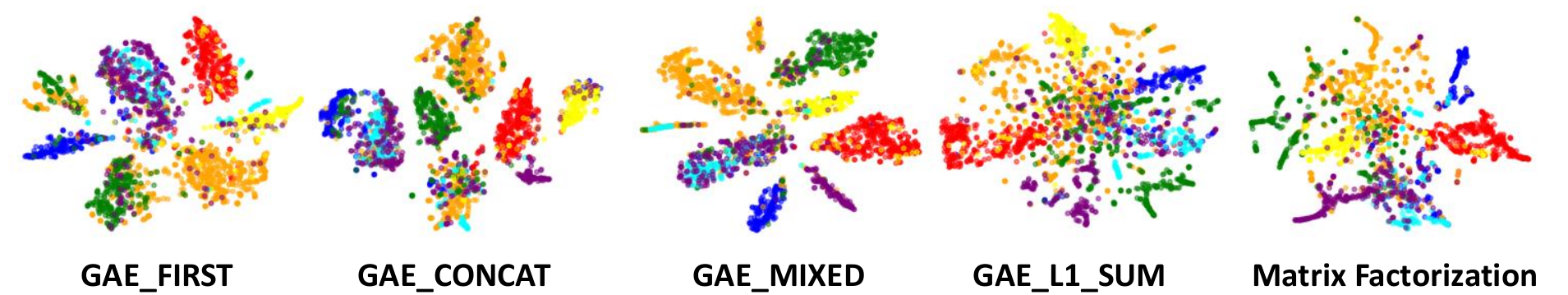}}
\caption{2D t-SNE projection of the embeddings of Cora. The embeddings are colored according to the seven different ground-truth labels of the dataset.
}
\label{fig:groundtruth}

 \vspace*{\floatsep}
 
\centerline{\includegraphics[scale=0.8]{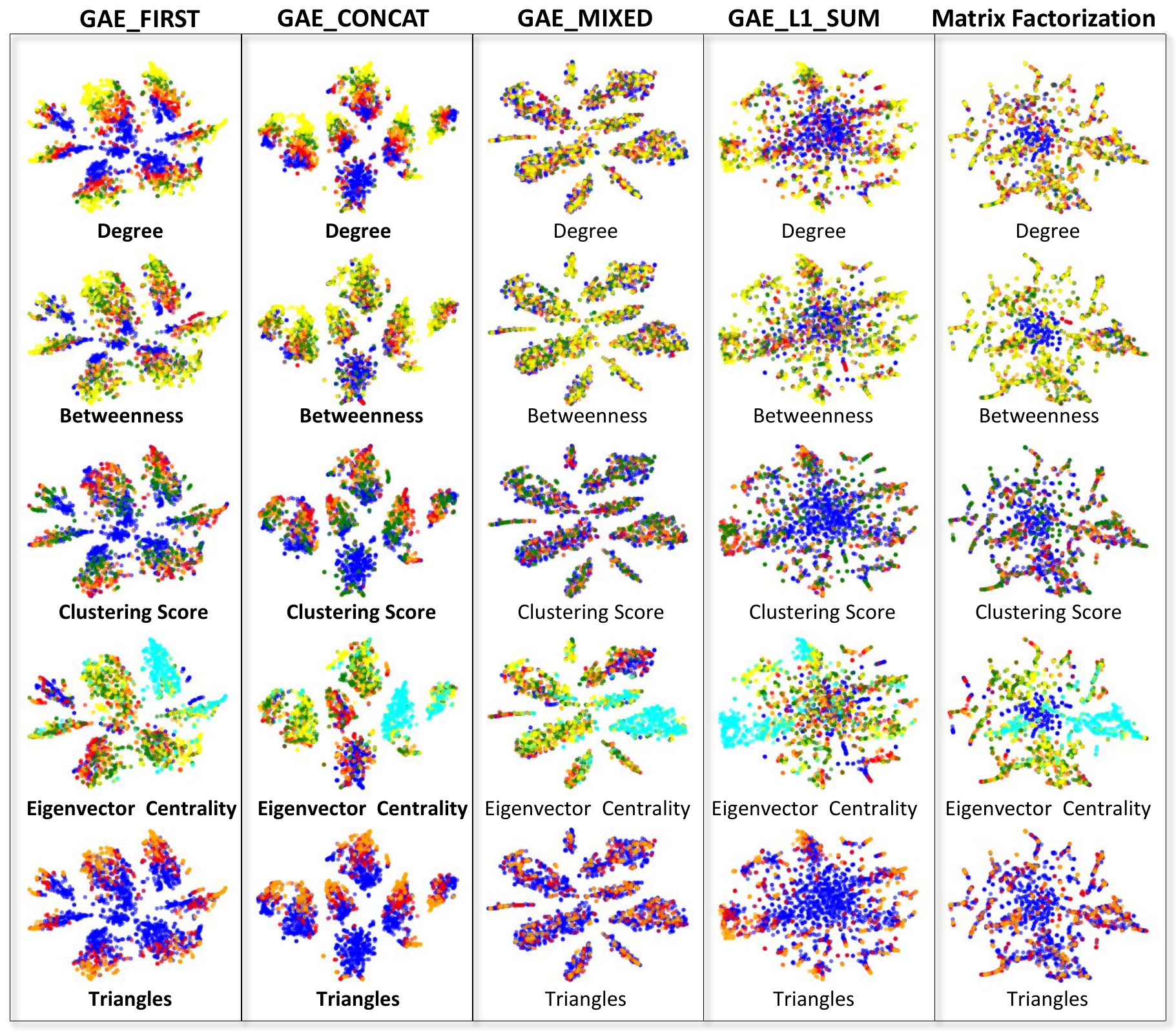}}
\caption{The embeddings of 5 models projected in 2D for the Cora dataset and colored according to the the topological classes of the vertices (Degree Class, Betweenness Class, Local Clustering Score Class, Eigenvector Centrality , Triangle Count). The colors listed in ascending order are: Blue, Red, Orange, Green, Yellow and Cyan. The figure clearly shows that the embeddings of GAE\_FIRST and GAE\_CONCAT (highlighted in bold) are organized according to the topological classes of the vertices.}
\label{fig:embeddingcolored}

\end{figure*}
\begin{figure*}[!t]
\centerline{\includegraphics[scale=0.5]{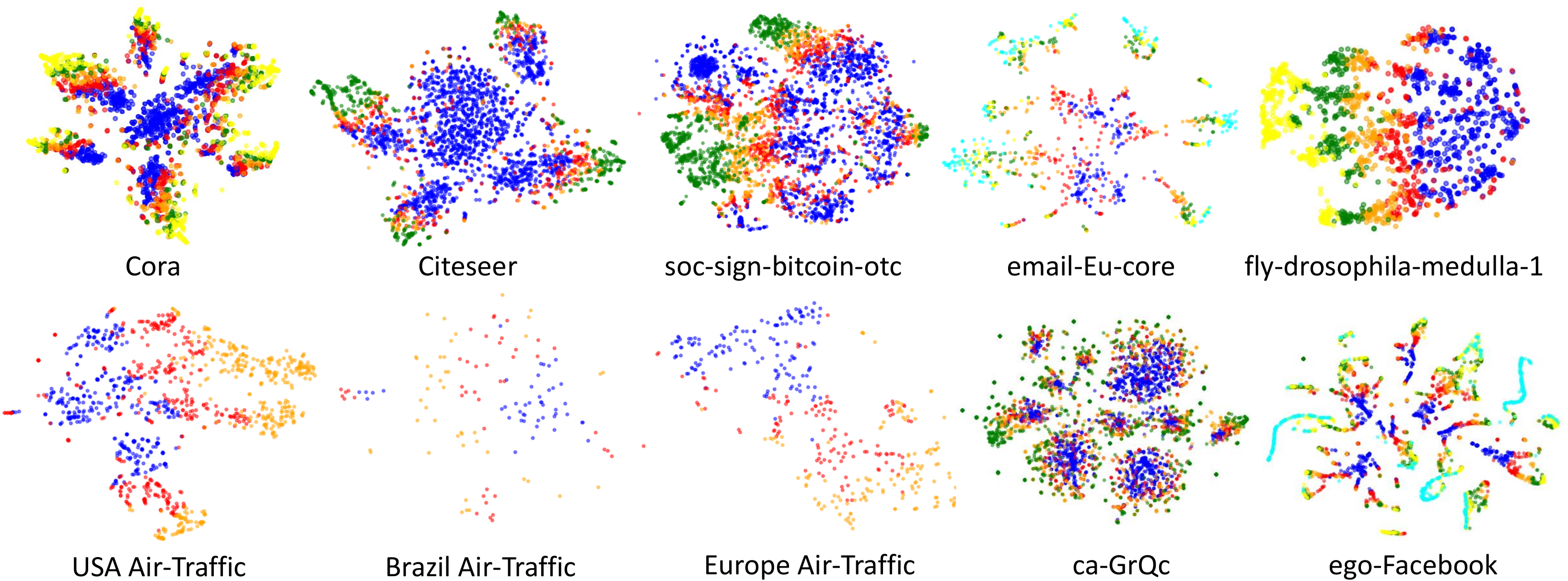}}
\caption{An illustration of the hierarchical arrangement of the node Degree classes for the embeddings generated by GAE\_FIRST. Every color represents a range of degrees as defined in the binning phase of the experiments. The plots show that the degrees in the embedding are arranged in a hierarchy from the smallest (Blue) to the largest (Cyan).}
\label{fig:hierarchy}
\end{figure*}

\subsection{Visualisation}

In this section, we aim to visualize  the embeddings for the purpose of better understanding the success of GAE\_FIRST and GAE\_CONCAT in preserving the topological features. To this end, we project the embeddings to two dimensions using t-SNE (t-distributed Stochastic Neighbor Embedding)\citep{Maaten2008}. Figures \ref{fig:groundtruth} and  \ref{fig:embeddingcolored} \footnote{All figures in this paper are best viewed in color/screen.} display the 2D projections of the embeddings generated on Cora dataset. As we can see from Figure \ref{fig:groundtruth}, when we color the nodes  according to the ground-truth, we notice that, for GAE\_FIRST and GAE\_CONCAT, each cluster of points in the plot is dominated by a class. This is due to the fact that the embedding models are mostly optimized to preserve the first-order and second-order proximity which are generally correlated with the ground-truth labels of the nodes. For example, in the case of the Cora dataset, the nodes are articles, the edges are citations, and the labels are the field of publication per article. Two articles (nodes) that have a high first-order and second-order proximity tend to belong to the same field of publication (i.e., they have the same label). A model that well preserves the orders of proximity, will also be preserving the labels of the graph.

However, in Figure \ref{fig:embeddingcolored}, when we color the nodes according to their topological feature classes (Degree class, Local Clustering class, Betweenness class,  Eigenvector Centrality class, and Triangle Count class), we notice that, for GAE\_FIRST and GAE\_CONCAT, the colors are clearly arranged in an organized manner, in contrast to the other models. In fact, the colors for the Degree in particular, exhibit a hierarchy in their arrangement, where, the nodes that have the lowest degree are projected close together (colored blue), followed by the nodes with higher degree (colored red) in an ascending order, all the way to the nodes belonging to the highest degree range (colored yellow). For GAE\_FIRST in particular, this hierarchy in the arrangement of the degrees is consistent with the majority of the tested datasets as illustrated by Figure \ref{fig:hierarchy}. 

This neat arrangement in the topological features is, however, absent from the embeddings of the other models. Consider, for example, the models that use the MEAN aggregation rule such as GAE\_MIXED. As we can see from Figure \ref{fig:embeddingcolored}, the topological features are arbitrarily distributed without any structure in the embeddings. The pictorial illustration of the organization of the topological classes in the embeddings of GAE\_FIRST and GAE\_CONCAT corroborate our claim on the fact that these two models preserve the topological features in their embeddings. Furthermore, we can confirm the failure of GAE\_L1\_SUM in capturing the topological features on Cora, as it failed to preserve the second-order proximity  (GAE\_L1\_SUM in Figure \ref{fig:groundtruth} shows no clear separation of ground-truth labels) and therefore did not preserve the topological features  (GAE\_L1\_SUM in Figure  \ref{fig:embeddingcolored})

\begin{table}[!ht]
\hspace*{-1.5cm}
\centering
\fontsize{6}{6}\selectfont
\setlength{\tabcolsep}{2pt}
\begin{tabular}{lcccccc}
\hline\hline
 & \multicolumn{3}{c}{\textbf{Cora}} & \multicolumn{3}{c}{\textbf{email-Eu-core}} \\ \cline{2-7} 
Models & DB & CH & SC & DB & CH & SC \\ \hline
GAE\_FIRST & 3.095 ±0.288 & 144.535 ±15.695 & \multicolumn{1}{c|}{0.066 ±0.015} & 3.965 ±0.214 & 15.863 ±1.547 & 0.026 ±0.006 \\
GAE\_CONCAT & 3.248 ±0.523 & 213.227 ±26.267 & \multicolumn{1}{c|}{0.100 ±0.033} & 4.204 ±0.268 & 11.018 ±1.077 & 0.043 ±0.005 \\
GAE\_L1\_SUM & 3.899 ±0.509 & 92.613 ±33.691 & \multicolumn{1}{c|}{-0.065 ±0.028} & 4.764 ±0.35 & 6.472 ±0.774 & -0.008 ±0.008 \\
GAE\_L2\_SUM & 2.938 ±0.371 & 297.779 ±31.627 & \multicolumn{1}{c|}{0.176 ±0.029} & 4.687 ±0.386 & 5.095 ±0.446 & -0.047 ±0.01 \\
GAE\_MEAN & 2.158 ±0.518 & \textbf{392.946 ±59.420} & \multicolumn{1}{c|}{0.186 ±0.024} & 6.205 ±0.202 & \textbf{3.545 ±0.244} & -0.017 ±0.004 \\
GAE\_MIXED & \textbf{2.051 ±0.401} & 390.704 ±48.115 & \multicolumn{1}{c|}{\textbf{0.211 ±0.027}} & 6.043 ±0.142 & 3.188 ±0.107 & \textbf{-0.038 ±0.004} \\
GAE\_SPECTRAL & 5.000 ±0.429 & 40.336 ±5.400 & \multicolumn{1}{c|}{-0.044 ±0.010} & 6.165 ±0.131 & 3.728 ±0.3 & 0.01 ±0.005 \\
Matrix Factorization & 5.174 ±0.085 & 10.706 ±0.998 & \multicolumn{1}{c|}{-0.196 ±0.035} & \textbf{7.353 ±0.13} & 0.871 ±0.03 & -0.135 ±0.001 \\
Node2Vec-S & 29.233 ±1.473 & 1.026 ±0.127 & \multicolumn{1}{c|}{-0.034 ±0.003} & 11.506 ±3.325 & 1.021 ±0.345 & -0.059 ±0.01 \\
Node2Vec-H & 28.532 ±1.293 & 1.074 ±0.092 & \multicolumn{1}{c|}{-0.042 ±0.003} & 11.635 ±2.015 & 0.764 ±0.155 & -0.07 ±0.012 \\ \hline
 & \multicolumn{1}{l}{} & \multicolumn{1}{l}{} & \multicolumn{1}{l}{} & \multicolumn{1}{l}{} & \multicolumn{1}{l}{} & \multicolumn{1}{l}{} \\
 & \multicolumn{3}{c}{\textbf{Citeseer}} & \multicolumn{3}{c}{\textbf{USA Air-Traffic}} \\ \cline{2-7} 
Models & DB & CH & SC & DB & CH & SC \\ \hline
GAE\_FIRST & 4.294 ±0.594 & 131.461 ±14.800 & \multicolumn{1}{c|}{-0.061 ±0.016} & \textbf{4.647 ±0.310} & \textbf{139.871 ±9.416} & 0.019 ±0.007 \\
GAE\_CONCAT & 4.721 ±0.606 & 155.863 ±20.255 & \multicolumn{1}{c|}{-0.026 ±0.019} & 5.906 ±0.419 & 93.810 ±8.284 & \textbf{0.044 ±0.009} \\
GAE\_L1\_SUM & 6.366 ±1.432 & 34.739 ±4.089 & \multicolumn{1}{c|}{-0.215 ±0.022} & 4.946 ±0.402 & 57.761 ±5.151 & -0.050 ±0.010 \\
GAE\_L2\_SUM & 5.612 ±1.048 & 209.234 ±22.836 & \multicolumn{1}{c|}{0.014 ±0.032} & 8.219 ±1.201 & 45.644 ±6.358 & 0.016 ±0.008 \\
GAE\_MEAN & 4.088 ±0.608 & 306.785 ±71.098 & \multicolumn{1}{c|}{0.094 ±0.041} & 7.624 ±0.686 & 31.904 ±2.373 & 0.016 ±0.002 \\
GAE\_MIXED & \textbf{3.151 ±0.378} & \textbf{371.070 ±55.394} & \multicolumn{1}{c|}{\textbf{0.172 ±0.029}} & 8.843 ±0.733 & 42.489 ±3.717 & 0.025 ±0.003 \\
GAE\_SPECTRAL & 8.882 ±0.662 & 26.218 ±2.863 & \multicolumn{1}{c|}{-0.032 ±0.005} & 12.555 ±1.235 & 6.217 ±0.995 & -0.088 ±0.003 \\
Matrix Factorization & 9.615 ±0.265 & 10.407 ±0.171 & \multicolumn{1}{c|}{-0.07 ±0.005} & 9.781 ±0.029 & 2.705 ±0.017 & -0.183 ±0.001 \\
Node2Vec-S & 33.581 ±1.514 & 1.087 ±0.115 & \multicolumn{1}{c|}{-0.01 ±0.002} & 26.090 ±1.715 & 0.952 ±0.165 & -0.033 ±0.003 \\
Node2Vec-H & 34.023 ±1.832 & 1.042 ±0.119 & \multicolumn{1}{c|}{-0.012 ±0.002} & 25.775 ±2.798 & 1.097 ±0.293 & -0.028 ±0.003 \\ \hline
 & \multicolumn{1}{l}{} & \multicolumn{1}{l}{} & \multicolumn{1}{l}{} & \multicolumn{1}{l}{} & \multicolumn{1}{l}{} & \multicolumn{1}{l}{} \\
 & \multicolumn{3}{c}{\textbf{Europe Air-Traffic}} & \multicolumn{3}{c}{\textbf{Brazil Air-Traffic}} \\ \cline{2-7} 
Models & DB & CH & SC & DB & CH & SC \\ \hline
GAE\_FIRST & \textbf{5.347 ±0.351} & \textbf{34.376 ±3.991} & \multicolumn{1}{c|}{-0.015 ±0.003} & \textbf{3.965 ±0.214} & \textbf{15.863 ±1.547} & 0.026 ±0.006 \\
GAE\_CONCAT & 6.126 ±0.394 & 25.105 ±2.491 & \multicolumn{1}{c|}{\textbf{-0.010 ±0.004}} & 4.204 ±0.268 & 11.018 ±1.077 & \textbf{0.043 ±0.005} \\
GAE\_L1\_SUM & 6.938 ±0.208 & 11.394 ±0.402 & \multicolumn{1}{c|}{-0.061 ±0.003} & 4.764 ±0.350 & 6.472 ±0.774 & -0.008 ±0.008 \\
GAE\_L2\_SUM & 7.782 ±0.657 & 9.631 ±1.413 & \multicolumn{1}{c|}{-0.061 ±0.006} & 4.687 ±0.386 & 5.095 ±0.446 & -0.047 ±0.010 \\
GAE\_MEAN & 10.252 ±0.563 & 5.414 ±0.455 & \multicolumn{1}{c|}{-0.056 ±0.004} & 6.205 ±0.202 & 3.545 ±0.244 & -0.017 ±0.004 \\
GAE\_MIXED & 10.174 ±0.577 & 5.699 ±0.365 & \multicolumn{1}{c|}{-0.070 ±0.002} & 6.043 ±0.142 & 3.188 ±0.107 & -0.038 ±0.004 \\
GAE\_SPECTRAL & 8.158 ±0.240 & 12.623 ±1.197 & \multicolumn{1}{c|}{-0.010 ±0.001} & 6.165 ±0.131 & 3.728 ±0.300 & 0.010 ±0.005 \\
Matrix Factorization & 11.537 ±0.000 & 1.140 ±0.000 & \multicolumn{1}{c|}{-0.166 ±0.000} & 7.353 ±0.13 & 0.871 ±0.030 & -0.135 ±0.001 \\
Node2Vec-S & 11.287 ±1.993 & 4.281 ±0.728 & \multicolumn{1}{c|}{-0.033 ±0.004} & 11.506 ±3.325 & 1.021 ±0.345 & -0.059 ±0.010 \\
Node2Vec-H & 11.871 ±1.319 & 3.590 ±0.396 & \multicolumn{1}{c|}{-0.020 ±0.003} & 11.635 ±2.015 & 0.764 ±0.155 & -0.070 ±0.012 \\ \hline\hline
\end{tabular}
\caption {Embedding clusters homogeneity.}
\label{tab:clusterhomogeneityresults}
\end{table}

\subsection{\textbf{Task 1}:Embedding Clusters Homogeneity}

Next, we study the homogeneity of the embeddings when clustered according to the ground truth labels of each dataset. We look into three different evaluation metrics. \textbf{Davies-Bouldin Index (DB)}: The lower the value the better the results. \textbf{Silhouette Score (SC)}: Score between -1 and 1 with 1 being the best score. \textbf{Calinski-Harabasz (CH)}: The higher the value, the better the consistency of the clusters. Table \ref{tab:clusterhomogeneityresults} reports the results on the datasets with ground truth labels.

\begin{figure}
\centerline{\includegraphics[scale=0.5]{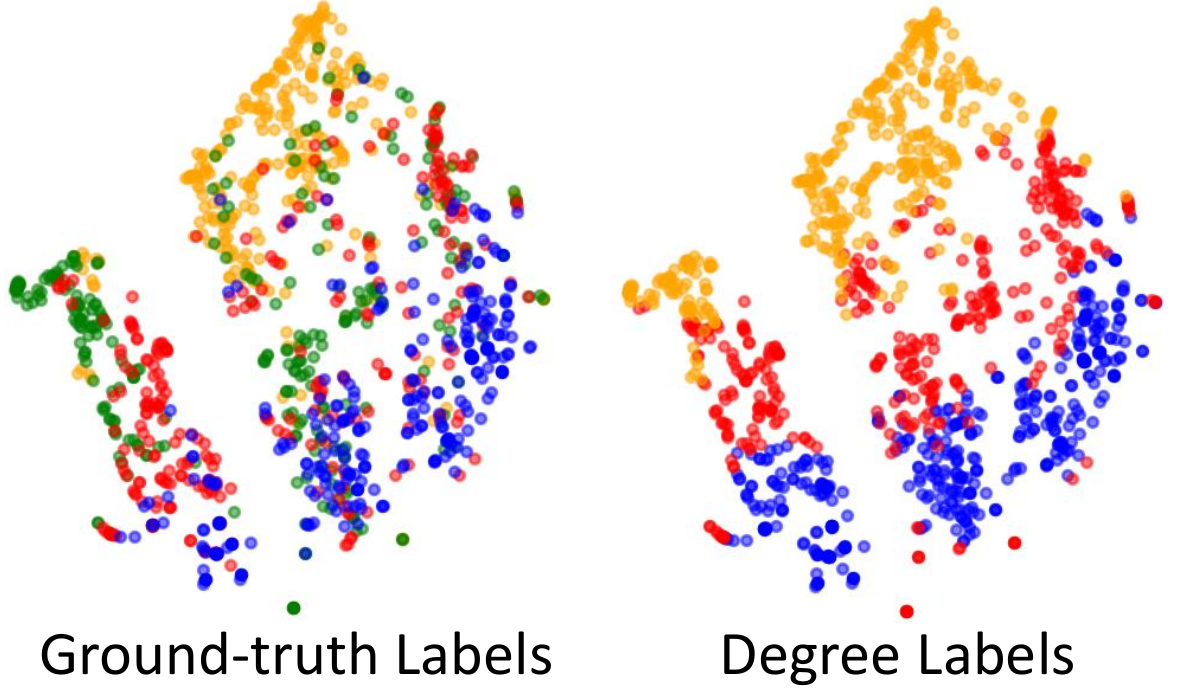}}
\caption{ 2D t-SNE projection of the GAE\_FIRST embeddings for USA Air-Traffic dataset - The colors show that, to a large extent, the vertices that belong to the same ground-truth class also belong to the same degree class.}
\label{fig:imageusa}
\vspace{-15pt}
\end{figure}

We find two trends in the results. On one hand, the models that use the mean and spectral aggregation rules (GAE\_MIXED, GAE\_MEAN and GAE\_SPECTRAL) perform best on the Cora, Citeseer and email-Eu-core datasets. This is most probably due to the smoothing effect that these two rules have on the attributes \citep{Li2018}, leading to more homogeneous embeddings. We also notice that GAE\_MIXED, the model that reconstructs two orders of proximity, gave the best results on two out of the three datasets.  On the other hand, the models that preserved the topological features (GAE\_FIRST and GAE\_CONCAT) dominated the three flight datasets (USA, Europe, and Brazil). The nodes in these datasets are airports, the edges are flights connecting the airports and the ground-truth labels are set according to the amount of activity that each airport receives. Therefore, there is a direct relation between the degree of each node (the number of flights that connect to it) and its label (the amount of activity the airport receives). The knowledgeable reader can observe this fact from Figure \ref{fig:imageusa}, where we notice that, to a large extent, the nodes that have the same ground-truth label also belong to the same degree class. That is why the models that capture the degree in the embedding best perform on these datasets.

\subsection{\textbf{Task 2}: Node Clustering}

\begin{table*}[!t]
\hspace*{-1.5cm}
\centering
\fontsize{6}{6}\selectfont
\setlength{\tabcolsep}{1.5 pt}
\begin{tabular}{lcccccc}
\hline\hline
 & \multicolumn{3}{c}{\textbf{Cora}} & \multicolumn{3}{c}{\textbf{email-Eu-core}} \\ \cline{2-7} 
Models & ACC & NMI & ARI & ACC & NMI & ARI \\ \hline
GAE\_FIRST & 0.438 ±0.039 & 0.324 ±0.030 & \multicolumn{1}{c|}{0.133 ±0.027} & 0.363 ±0.015 & 0.564 ±0.01 & 0.228 ±0.016 \\
GAE\_CONCAT & 0.586 ±0.059 & 0.440 ±0.031 & \multicolumn{1}{c|}{0.367 ±0.059} & 0.344 ±0.018 & 0.558 ±0.012 & 0.247 ±0.025 \\
GAE\_L1\_SUM & 0.372 ±0.018 & 0.194 ±0.029 & \multicolumn{1}{c|}{0.044 ±0.016} & 0.364 ±0.008 & 0.531 ±0.008 & 0.163 ±0.013 \\
GAE\_L2\_SUM & 0.629 ±0.036 & 0.469 ±0.022 & \multicolumn{1}{c|}{0.403 ±0.028} & 0.384 ±0.017 & 0.567 ±0.011 & 0.327 ±0.028 \\
GAE\_MEAN & \textbf{0.668 ±0.021} & 0.505 ±0.021 & \multicolumn{1}{c|}{\textbf{0.430 ±0.031}} & 0.432 ±0.011 & \textbf{0.605 ±0.008} & \textbf{0.356 ±0.011} \\
GAE\_MIXED & 0.659 ±0.019 & \textbf{0.507 ±0.017} & \multicolumn{1}{c|}{0.417 ±0.028} & \textbf{0.438 ±0.035} & 0.603 ±0.013 & 0.350 ±0.038 \\
GAE\_SPECTRAL & 0.419 ±0.050 & 0.231 ±0.031 & \multicolumn{1}{c|}{0.158 ±0.038} & 0.395 ±0.040 & 0.547 ±0.057 & 0.172 ±0.013 \\
Matrix Factorization & 0.304 ±0.002 & 0.010 ±0.002 & \multicolumn{1}{c|}{0.000 ±0.001} & 0.170 ±0.020 & 0.218 ±0.032 & 0.006 ±0.007 \\
Node2Vec-S & 0.229 ±0.006 & 0.005 ±0.001 & \multicolumn{1}{c|}{0.002 ±0.002} & 0.109 ±0.003 & 0.195 ±0.004 & 0.000 ±0.001 \\
Node2Vec-H & 0.217 ±0.005 & 0.004 ±0.001 & \multicolumn{1}{c|}{0.001 ±0.003} & 0.109 ±0.004 & 0.183 ±0.005 & 0.000 ±0.001 \\ \hline
 & \multicolumn{1}{l}{} & \multicolumn{1}{l}{} & \multicolumn{1}{l}{} & \multicolumn{1}{l}{} & \multicolumn{1}{l}{} & \multicolumn{1}{l}{} \\
 & \multicolumn{3}{c}{\textbf{Citeseer}} & \multicolumn{3}{c}{\textbf{USA Air-Traffic}} \\ \cline{2-7} 
Models & ACC & NMI & ARI & ACC & NMI & ARI \\ \hline
GAE\_FIRST & 0.370 ±0.027 & 0.181 ±0.027 & \multicolumn{1}{c|}{0.048 ±0.011} & 0.468 ±0.009 & \textbf{0.273 ±0.005} & \textbf{0.196 ±0.004} \\
GAE\_CONCAT & 0.450 ±0.035 & 0.217 ±0.024 & \multicolumn{1}{c|}{0.156 ±0.029} & \textbf{0.484 ±0.01} & 0.200 ±0.012 & 0.184 ±0.016 \\
GAE\_L1\_SUM & 0.231 ±0.007 & 0.056 ±0.009 & \multicolumn{1}{c|}{-0.001 ±0.001} & 0.468 ±0.016 & 0.224 ±0.028 & 0.169 ±0.024 \\
GAE\_L2\_SUM & 0.460 ±0.037 & 0.234 ±0.028 & \multicolumn{1}{c|}{0.178 ±0.031} & 0.452 ±0.021 & 0.141 ±0.025 & 0.134 ±0.028 \\
GAE\_MEAN & 0.565 ±0.041 & 0.309 ±0.030 & \multicolumn{1}{c|}{0.297 ±0.036} & 0.432 ±0.038 & 0.125 ±0.025 & 0.120 ±0.030 \\
GAE\_MIXED & \textbf{0.609 ±0.037} & \textbf{0.361 ±0.029} & \multicolumn{1}{c|}{\textbf{0.352 ±0.039}} & 0.469 ±0.007 & 0.182 ±0.024 & 0.184 ±0.024 \\
GAE\_SPECTRAL & 0.353 ±0.038 & 0.098 ±0.019 & \multicolumn{1}{c|}{0.089 ±0.020} & 0.351 ±0.019 & 0.059 ±0.012 & 0.033 ±0.011 \\
Matrix Factorization & 0.224 ±0.018 & 0.024 ±0.024 & \multicolumn{1}{c|}{0.004 ±0.006} & 0.254 ±0.004 & 0.01 ±0.003 & 0.000 ±0.000 \\
Node2Vec-S & 0.204 ±0.003 & 0.003 ±0.001 & \multicolumn{1}{c|}{0.000 ±0.001} & 0.282 ±0.007 & 0.006 ±0.002 & 0.003 ±0.001 \\
Node2Vec-H & 0.204 ±0.004 & 0.003 ±0.001 & \multicolumn{1}{c|}{-0.000 ±0.001} & 0.287 ±0.007 & 0.008 ±0.003 & 0.005 ±0.003 \\ \hline
 & \multicolumn{1}{l}{} & \multicolumn{1}{l}{} & \multicolumn{1}{l}{} & \multicolumn{1}{l}{} & \multicolumn{1}{l}{} & \multicolumn{1}{l}{} \\
 & \multicolumn{3}{c}{\textbf{Europe Air-Traffic}} & \multicolumn{3}{c}{\textbf{Brazil Air-Traffic}} \\ \cline{2-7} 
Models & DB & CH & SC & DB & CH & SC \\ \hline
GAE\_FIRST & 0.425 ±0.010 & 0.199 ±0.017 & \multicolumn{1}{c|}{0.175 ±0.015} & 0.479 ±0.021 & \textbf{0.304 ±0.009} & 0.219 ±0.012 \\
GAE\_CONCAT & 0.394 ±0.018 & 0.108 ±0.012 & \multicolumn{1}{c|}{0.098 ±0.017} & 0.490 ±0.016 & 0.293 ±0.025 & \textbf{0.235 ±0.022} \\
GAE\_L1\_SUM & 0.411 ±0.011 & 0.133 ±0.006 & \multicolumn{1}{c|}{0.123 ±0.005} & \textbf{0.496 ±0.042} & 0.257 ±0.048 & 0.187 ±0.046 \\
GAE\_L2\_SUM & 0.362 ±0.023 & 0.078 ±0.021 & \multicolumn{1}{c|}{0.051 ±0.020} & 0.420 ±0.028 & 0.175 ±0.041 & 0.090 ±0.041 \\
GAE\_MEAN & 0.336 ±0.010 & 0.061 ±0.012 & \multicolumn{1}{c|}{0.027 ±0.008} & 0.351 ±0.014 & 0.079 ±0.015 & 0.022 ±0.012 \\
GAE\_MIXED & 0.329 ±0.011 & 0.058 ±0.011 & \multicolumn{1}{c|}{0.020 ±0.008} & 0.354 ±0.023 & 0.107 ±0.023 & 0.029 ±0.017 \\
GAE\_SPECTRAL & \textbf{0.439 ±0.016} & \textbf{0.213 ±0.010} & \multicolumn{1}{c|}{\textbf{0.178 ±0.011}} & 0.482 ±0.031 & 0.278 ±0.050 & 0.215 ±0.038 \\
Matrix Factorization & 0.261 ±0.011 & 0.025 ±0.011 & \multicolumn{1}{c|}{0.001 ±0.001} & 0.257 ±0.005 & 0.043 ±0.005 & -0.001 ±0.000 \\
Node2Vec-S & 0.316 ±0.013 & 0.026 ±0.009 & \multicolumn{1}{c|}{0.017 ±0.008} & 0.305 ±0.013 & 0.020 ±0.007 & -0.003 ±0.007 \\
Node2Vec-H & 0.320 ±0.007 & 0.029 ±0.005 & \multicolumn{1}{c|}{0.014 ±0.004} & 0.310 ±0.018 & 0.022 ±0.009 & 0.000 ±0.007 \\ \hline\hline
\end{tabular}
\caption{Clustering node embeddings - using K-means.}
\label{tab:clusteringkmeans}
\end{table*}

\begin{table*}[!t]
\hspace*{-1.5cm}
\centering
\fontsize{6}{6}\selectfont
\setlength{\tabcolsep}{1.5 pt}
\begin{tabular}{lcccccc}
\hline\hline
 & \multicolumn{3}{c}{\textbf{Cora}} & \multicolumn{3}{c}{\textbf{email-Eu-core}} \\ \cline{2-7} 
Models & ACC & NMI & ARI & ACC & NMI & ARI \\ \hline
GAE\_FIRST & 0.555 ±0.042 & 0.439 ±0.030 & \multicolumn{1}{c|}{0.342 ±0.047} & 0.305 ±0.009 & 0.443 ±0.012 & 0.164 ±0.014 \\
GAE\_CONCAT & 0.580 ±0.064 & 0.435 ±0.036 & \multicolumn{1}{c|}{0.362 ±0.060} & 0.330 ±0.018 & 0.482 ±0.014 & 0.232 ±0.019 \\
GAE\_L1\_SUM & 0.397 ±0.056 & 0.234 ±0.092 & \multicolumn{1}{c|}{0.070 ±0.051} & 0.294 ±0.011 & 0.428 ±0.006 & 0.149 ±0.015 \\
GAE\_L2\_SUM & 0.624 ±0.044 & 0.464 ±0.027 & \multicolumn{1}{c|}{0.397 ±0.037} & 0.338 ±0.009 & 0.490 ±0.010 & 0.243 ±0.007 \\
GAE\_MEAN & \textbf{0.656 ±0.035} & \textbf{0.502 ±0.021} & \multicolumn{1}{c|}{\textbf{0.423 ±0.029}} & \textbf{0.369 ±0.011} & \textbf{0.508 ±0.018} & \textbf{0.262 ±0.023} \\
GAE\_MIXED & 0.638 ±0.030 & 0.492 ±0.019 & \multicolumn{1}{c|}{0.402 ±0.029} & 0.359 ±0.012 & 0.497 ±0.015 & 0.215 ±0.023 \\
GAE\_SPECTRAL & 0.309 ±0.033 & 0.048 ±0.057 & \multicolumn{1}{c|}{0.006 ±0.032} & 0.272 ±0.027 & 0.312 ±0.041 & 0.042 ±0.025 \\
Matrix Factorization & 0.293 ±0.003 & 0.036 ±0.007 & \multicolumn{1}{c|}{-0.003 ±0.003} & 0.245 ±0.017 & 0.387 ±0.010 & 0.153 ±0.015 \\
Node2Vec-S & 0.250 ±0.010 & 0.004 ±0.001 & \multicolumn{1}{c|}{-0.001 ±0.003} & 0.100 ±0.004 & 0.052 ±0.005 & -0.000 ±0.001 \\
Node2Vec-H & 0.202 ±0.015 & 0.004 ±0.001 & \multicolumn{1}{c|}{-0.000 ±0.001} & 0.099 ±0.004 & 0.052 ±0.005 & -0.001 ±0.002 \\ \hline
 & \multicolumn{1}{l}{} & \multicolumn{1}{l}{} & \multicolumn{1}{l}{} & \multicolumn{1}{l}{} & \multicolumn{1}{l}{} & \multicolumn{1}{l}{} \\
 & \multicolumn{3}{c}{\textbf{Citeseer}} & \multicolumn{3}{c}{\textbf{USA Air-Traffic}} \\ \cline{2-7} 
Models & ACC & NMI & ARI & ACC & NMI & ARI \\ \hline
GAE\_FIRST & 0.467 ±0.046 & 0.240 ±0.038 & \multicolumn{1}{c|}{0.172 ±0.039} & \textbf{0.458 ±0.056} & 0.232 ±0.022 & \textbf{0.210 ±0.040} \\
GAE\_CONCAT & 0.468 ±0.036 & 0.223 ±0.022 & \multicolumn{1}{c|}{0.176 ±0.029} & 0.410 ±0.048 & 0.202 ±0.022 & 0.162 ±0.035 \\
GAE\_L1\_SUM & 0.259 ±0.035 & 0.060 ±0.032 & \multicolumn{1}{c|}{0.017 ±0.023} & 0.413 ±0.024 & \textbf{0.238 ±0.021} & 0.160 ±0.014 \\
GAE\_L2\_SUM & 0.463 ±0.041 & 0.236 ±0.028 & \multicolumn{1}{c|}{0.182 ±0.029} & 0.382 ±0.036 & 0.158 ±0.009 & 0.122 ±0.018 \\
GAE\_MEAN & 0.558 ±0.047 & 0.309 ±0.029 & \multicolumn{1}{c|}{0.295 ±0.040} & 0.348 ±0.023 & 0.171 ±0.011 & 0.109 ±0.013 \\
GAE\_MIXED & \textbf{0.601 ±0.037} & \textbf{0.368 ±0.021} & \multicolumn{1}{c|}{\textbf{0.360 ±0.028}} & 0.369 ±0.015 & 0.196 ±0.008 & 0.118 ±0.009 \\
GAE\_SPECTRAL & 0.211 ±0.004 & 0.016 ±0.005 & \multicolumn{1}{c|}{0.000 ±0.001} & 0.354 ±0.030 & 0.071 ±0.020 & 0.062 ±0.029 \\
Matrix Factorization & 0.427 ±0.017 & 0.170 ±0.010 & \multicolumn{1}{c|}{0.158 ±0.011} & 0.309 ±0.021 & 0.063 ±0.013 & 0.046 ±0.013 \\
Node2Vec-S & 0.211 ±0.003 & 0.004 ±0.001 & \multicolumn{1}{c|}{0.001 ±0.001} & 0.232 ±0.012 & 0.005 ±0.001 & 0.000 ±0.001 \\
Node2Vec-H & 0.206 ±0.003 & 0.004 ±0.001 & \multicolumn{1}{c|}{0.002 ±0.002} & 0.245 ±0.019 & 0.007 ±0.002 & 0.001 ±0.001 \\ \hline
 & \multicolumn{1}{l}{} & \multicolumn{1}{l}{} & \multicolumn{1}{l}{} & \multicolumn{1}{l}{} & \multicolumn{1}{l}{} & \multicolumn{1}{l}{} \\
 & \multicolumn{3}{c}{\textbf{Europe Air-Traffic}} & \multicolumn{3}{c}{\textbf{Brazil Air-Traffic}} \\ \cline{2-7} 
Models & ACC & NMI & ARI & ACC & NMI & ARI \\ \hline
GAE\_FIRST & \textbf{0.36 ±0.039} & \textbf{0.154 ±0.034} & \multicolumn{1}{c|}{\textbf{0.120 ±0.034}} & 0.402 ±0.044 & \textbf{0.262 ±0.03} & 0.173 ±0.037 \\
GAE\_CONCAT & 0.321 ±0.027 & 0.123 ±0.013 & \multicolumn{1}{c|}{0.094 ±0.018} & 0.415 ±0.048 & 0.261 ±0.021 & 0.182 ±0.032 \\
GAE\_L1\_SUM & 0.343 ±0.014 & 0.110 ±0.012 & \multicolumn{1}{c|}{0.103 ±0.011} & \textbf{0.424 ±0.05} & 0.245 ±0.054 & \textbf{0.188 ±0.065} \\
GAE\_L2\_SUM & 0.332 ±0.035 & 0.123 ±0.016 & \multicolumn{1}{c|}{0.093 ±0.019} & 0.391 ±0.034 & 0.250 ±0.024 & 0.164 ±0.028 \\
GAE\_MEAN & 0.267 ±0.015 & 0.049 ±0.008 & \multicolumn{1}{c|}{0.031 ±0.009} & 0.330 ±0.020 & 0.166 ±0.023 & 0.090 ±0.018 \\
GAE\_MIXED & 0.266 ±0.016 & 0.048 ±0.013 & \multicolumn{1}{c|}{0.030 ±0.012} & 0.337 ±0.021 & 0.162 ±0.014 & 0.095 ±0.015 \\
GAE\_SPECTRAL & 0.280 ±0.013 & 0.073 ±0.019 & \multicolumn{1}{c|}{0.005 ±0.005} & 0.283 ±0.006 & 0.102 ±0.007 & 0.000 ±0.001 \\
Matrix Factorization & 0.238 ±0.000 & 0.022 ±0.000 & \multicolumn{1}{c|}{0.000 ±0.000} & 0.297 ±0.018 & 0.051 ±0.008 & 0.008 ±0.004 \\
Node2Vec-S & 0.233 ±0.012 & 0.030 ±0.006 & \multicolumn{1}{c|}{0.010 ±0.005} & 0.255 ±0.016 & 0.042 ±0.007 & -0.002 ±0.007 \\
Node2Vec-H & 0.234 ±0.012 & 0.035 ±0.005 & \multicolumn{1}{c|}{0.012 ±0.004} & 0.266 ±0.025 & 0.046 ±0.011 & 0.002 ±0.009 \\ \hline\hline
\end{tabular}
\caption{Clustering node embeddings - using FINCH.}
\label{tab:clusteringfinch}
\end{table*}

Let us now evaluate the suitability of the embeddings on the task  of node clustering. To this end, we apply K-means and a recently proposed algorithm named FINCH (a hierarchical agglomerative method) \citep{Sarfraz2019} on the embeddings of each model\footnote{We draw the attention of the reader to the fact that the goal of this paper is not to evaluate clustering algorithms but to evaluate the embeddings generated by various graph learning models for the task of clustering. We elected the use of K-means due to its efficiency and simplicity. We selected FINCH \citep{Sarfraz2019} because it is a recent efficient parameter-free clustering algorithm that does not require any parameters setting, including the number of clusters.}. 
For evaluation, we consider three metrics: Adjusted Rand Index (ARI), Normalized Mutual Information (NMI) and Clustering Accuracy (ACC). All the metrics are between 0 and 1 with 1 being the best result. We use the node labels as ground truth. As depicted by Tables \ref{tab:clusteringkmeans} and \ref{tab:clusteringfinch}, the results on this task are consistent with those of the embedding homogeneity, with GAE\_MEAN and GAE\_MIXED performing best on Cora, Citeseer and Email-Eu-Code. On the other hand, GAE\_FIRST and GAE\_CONCAT performed best on the flight datasets. However, it is important to note that the models that did preserve the topological features under-performed the vanilla model GAE\_L2\_SUM on three out of the six dataset for the task of clustering. 

\subsection{\textbf{Task 3}: Node Classification}
\begin{table}[!t]
\hspace*{-1cm}
\centering
\fontsize{6}{6}\selectfont
\setlength{\tabcolsep}{1.5pt}
\begin{tabular}{lcccc}
\hline\hline
 & \multicolumn{2}{c}{\textbf{Cora}} & \multicolumn{2}{c}{\textbf{email-Eu-core}} \\ \cline{2-5} 
Models &  Macro-F1 & Micro-F1 &  Macro-F1 & Micro-F1 \\ \hline
GAE\_FIRST & \textbf{0.796 ±0.009} & \multicolumn{1}{c|}{\textbf{0.809 ±0.008}} & 0.516 ±0.018 & 0.683 ±0.010 \\
GAE\_CONCAT & 0.758 ±0.015 & \multicolumn{1}{c|}{0.780 ±0.014} & 0.447 ±0.018 & 0.639 ±0.012 \\
GAE\_L1\_SUM & 0.132 ±0.024 & \multicolumn{1}{c|}{0.348 ±0.022} & 0.475 ±0.011 & 0.629 ±0.006 \\
GAE\_L2\_SUM & 0.733 ±0.025 & \multicolumn{1}{c|}{0.76 ±0.016} & 0.330 ±0.016 & 0.544 ±0.009 \\
GAE\_MEAN & 0.765 ±0.016 & \multicolumn{1}{c|}{0.783 ±0.017} & 0.509 ±0.016 & 0.681 ±0.008 \\
GAE\_MIXED & 0.752 ±0.016 & \multicolumn{1}{c|}{0.775 ±0.015} & \textbf{0.532 ±0.014} & \textbf{0.696 ±0.007} \\
GAE\_SPECTRAL & 0.066 ±0.000 & \multicolumn{1}{c|}{0.302 ±0.000} & 0.025 ±0.001 & 0.169 ±0.002 \\
Matrix Factorization & 0.067 ±0.001 & \multicolumn{1}{c|}{0.303 ±0.001} & 0.008 ±0.000 & 0.112 ±0.000 \\
Node2Vec-S & 0.107 ±0.005 & \multicolumn{1}{c|}{0.259 ±0.004} & 0.024 ±0.006 & 0.077 ±0.007 \\
Node2Vec-H & 0.105 ±0.005 & \multicolumn{1}{c|}{0.257 ±0.004} & 0.020 ±0.005 & 0.078 ±0.005 \\ \hline
 & \multicolumn{1}{l}{} & \multicolumn{1}{l}{} & \multicolumn{1}{l}{} & \multicolumn{1}{l}{} \\
 & \multicolumn{2}{c}{\textbf{Citeseer}} & \multicolumn{2}{c}{\textbf{USA Air-Traffic}} \\ \cline{2-5} 
Models &  Macro-F1 & Micro-F1 &  Macro-F1 & Micro-F1 \\ \hline
GAE\_FIRST & \textbf{0.630 ±0.010} & \multicolumn{1}{c|}{0.682 ±0.011} & \textbf{0.650 ±0.009} & \textbf{0.654 ±0.009} \\
GAE\_CONCAT & 0.581 ±0.013 & \multicolumn{1}{c|}{0.634 ±0.016} & 0.643 ±0.009 & 0.646 ±0.010 \\
GAE\_L1\_SUM & 0.133 ±0.022 & \multicolumn{1}{c|}{0.264 ±0.019} & 0.634 ±0.022 & 0.634 ±0.022 \\
GAE\_L2\_SUM & 0.549 ±0.014 & \multicolumn{1}{c|}{0.614 ±0.017} & 0.552 ±0.031 & 0.564 ±0.027 \\
GAE\_MEAN & 0.584 ±0.017 & \multicolumn{1}{c|}{0.655 ±0.018} & 0.594 ±0.014 & 0.597 ±0.014 \\
GAE\_MIXED & 0.608 ±0.012 & \multicolumn{1}{c|}{\textbf{0.689 ±0.009}} & 0.593 ±0.014 & 0.597 ±0.012 \\
GAE\_SPECTRAL & 0.058 ±0.000 & \multicolumn{1}{c|}{0.211 ±0.000} & 0.414 ±0.021 & 0.439 ±0.019 \\
Matrix Factorization & 0.325 ±0.004 & \multicolumn{1}{c|}{0.433 ±0.003} & 0.482 ±0.001 & 0.499 ±0.001 \\
Node2Vec-S & 0.153 ±0.009 & \multicolumn{1}{c|}{0.189 ±0.010} & 0.239 ±0.014 & 0.248 ±0.011 \\
Node2Vec-H & 0.150 ±0.008 & \multicolumn{1}{c|}{0.189 ±0.008} & 0.240 ±0.016 & 0.247 ±0.015 \\ \hline
 & \multicolumn{1}{l}{} & \multicolumn{1}{l}{} & \multicolumn{1}{l}{} & \multicolumn{1}{l}{} \\
 & \multicolumn{2}{c}{\textbf{Europe Air-Traffic}} & \multicolumn{2}{c}{\textbf{Brazil Air-Traffic}} \\ \cline{2-5} 
Models & Macro-F1 & Micro-F1 &  Macro-F1 & Micro-F1 \\ \hline
GAE\_FIRST & \textbf{0.509 ±0.023} & \multicolumn{1}{c|}{\textbf{0.524 ±0.022}} & \textbf{0.564 ±0.032} & \textbf{0.572 ±0.030} \\
GAE\_CONCAT & 0.502 ±0.029 & \multicolumn{1}{c|}{0.518 ±0.029} & 0.554 ±0.029 & 0.569 ±0.030 \\
GAE\_L1\_SUM & 0.461 ±0.027 & \multicolumn{1}{c|}{0.474 ±0.026} & 0.522 ±0.018 & 0.528 ±0.022 \\
GAE\_L2\_SUM & 0.400 ±0.017 & \multicolumn{1}{c|}{0.414 ±0.021} & 0.488 ±0.023 & 0.509 ±0.024 \\
GAE\_MEAN & 0.468 ±0.014 & \multicolumn{1}{c|}{0.478 ±0.011} & 0.495 ±0.026 & 0.511 ±0.026 \\
GAE\_MIXED & 0.453 ±0.013 & \multicolumn{1}{c|}{0.469 ±0.013} & 0.496 ±0.023 & 0.514 ±0.022 \\
GAE\_SPECTRAL & 0.347 ±0.004 & \multicolumn{1}{c|}{0.469 ±0.004} & 0.290 ±0.0210 & 0.400 ±0.022 \\
Matrix Factorization & 0.149 ±0.000 & \multicolumn{1}{c|}{0.198 ±0.000} & 0.203 ±0.008 & 0.247 ±0.009 \\
Node2Vec-S & 0.311 ±0.020 & \multicolumn{1}{c|}{0.334 ±0.019} & 0.238 ±0.035 & 0.257 ±0.036 \\
Node2Vec-H & 0.297 ±0.029 & \multicolumn{1}{c|}{0.323 ±0.029} & 0.231 ±0.048 & 0.252 ±0.049 \\ \hline\hline
\end{tabular}
\caption{Node Classification - using LOGISTIC REGRESSION.}
\label{tab:classification}
\end{table}

Here, we evaluate the effect of the embeddings on the task of node classification. We apply the same models used for the classification of the topological features with the same parameters (LG-R, Linear SVM, SVM with RBF Kernel, and MLP). We use the embeddings as attributes and the dataset node labels as ground truth. We report the results of the Logistic Regression classifier (Table \ref{tab:classification}) evaluated with Micro-F1 and Macro-F1. Note that, as reported in the supplementary material website associated with this paper (\url{https://github.com/MH-0/RPGAE}), the results of the other models, that is, SVM-L, SVM-RBF and MLP, are to a large extent consistent with that of Logistic Regression.

From Table \ref{tab:classification}, we notice that GAE\_FIRST outperforms the other models on five out of the six datasets, with GAE\_CONCAT in close second, while GAE\_MIXED gave the best results on email-Eu-Core. The good performance of GAE\_FIRST and GAE\_CONCAT on the flight datasets for all three tasks is a reconfirmation for the importance of the encoding of the topological features in the embeddings, especially when the ground-truth labels are related to the structural role of the node. On the other hand, the consistent performance of GAE\_MIXED on all three tasks is a sign for the importance of the preservation of multiple orders of proximity in the embeddings.

\subsection{Analysis and Recommendations}
Following the extensive experiments that we have performed, we conclude that, though it is primarily beneficial to have the topological features encoded in the embeddings of the Graph Autoencoder, it is not always necessary. The choice of model should come down to two main factors: (1) The type of task for which the embeddings will be used, such as node classification or clustering, and (2) The correlation between the ground-truth labels of the dataset and the topological features of the nodes. For example, if the embeddings will be used for clustering, and the ground-truth labels are not related to any topological feature, we recommend using the MEAN or the SPECTRAL rule. This is so the model can harness their smoothing power, even if it will be at the expense of losing some of the topological information in the embeddings. On the other hand, if the ground-truth labels are correlated to the Degree of the node or its Triangle Count, we recommend using a model that employs the aggregation by SUM, and that includes the first layer of the encoder in the embedding. Furthermore, if the ground-truth labels are correlated to a feature that depends on the two-hop neighborhood such as Eigenvector Centrality or the Betweenness Centrality, we recommend concatenating the output of the first layers with the outputs of deeper layers to form the embedding. 

\section{Case Study}

In this section, we put our findings to work with an application that highlights the importance of adopting a “Topological Features Preservation” strategy when designing a GNN architecture.  The purpose of our case study is to show that by adopting an architecture capable of capturing topological features relevant to the underlying task of the model, we can boost the performance of this model without the need for any hand-crafted features. To illustrate our point, we have elected the use of DeepInf \citep{Qiu2018}, a recent deep learning approach for social influence prediction. In what follows, we lay out the scope of our case study over three main steps. First, we introduce DeepInf framework with a summary of its architecture, main components, and any modifications that we have made for the purpose of our case study. Next, we identify the topological features relevant to the social influence problem and adopt an architecture capable of capturing these features, following our findings and recommendations as suggested in the previous section. Finally, we conduct detailed experiments to assess the suitability of the adopted architecture.

\subsection{DeepInf Framework Summary}

DeepInf \citep{Qiu2018} is an end-to-end Graph Neural Network framework that predicts the user-level social influence in a graph. In other words, DeepInf tries to predict whether a user will perform a certain action (or be \textit{activated}) after being influenced by his surrounding or local network. For example, whether a user will buy a product that his friends bought or will share a story about a subject that his friends are sharing. When this problem is modeled as a graph, the status of a node (i.e., whether activated or not) is referred to as \textit{Action Status} $s_v^t\in \{0,1\}$ (1 for node $v$ activated at time $t$, 0 for node $v$ not activated at time $t$). DeepInf accomplishes the task of social influence prediction, by taking the structure and Action Status of the local neighborhood of a node as input and employing it to learn a latent feature representation of that node. The learned representation is subsequently used to predict whether that node will be activated or not in the future. 

\begin{figure*}[!t]
\centerline{\includegraphics[scale=0.75]{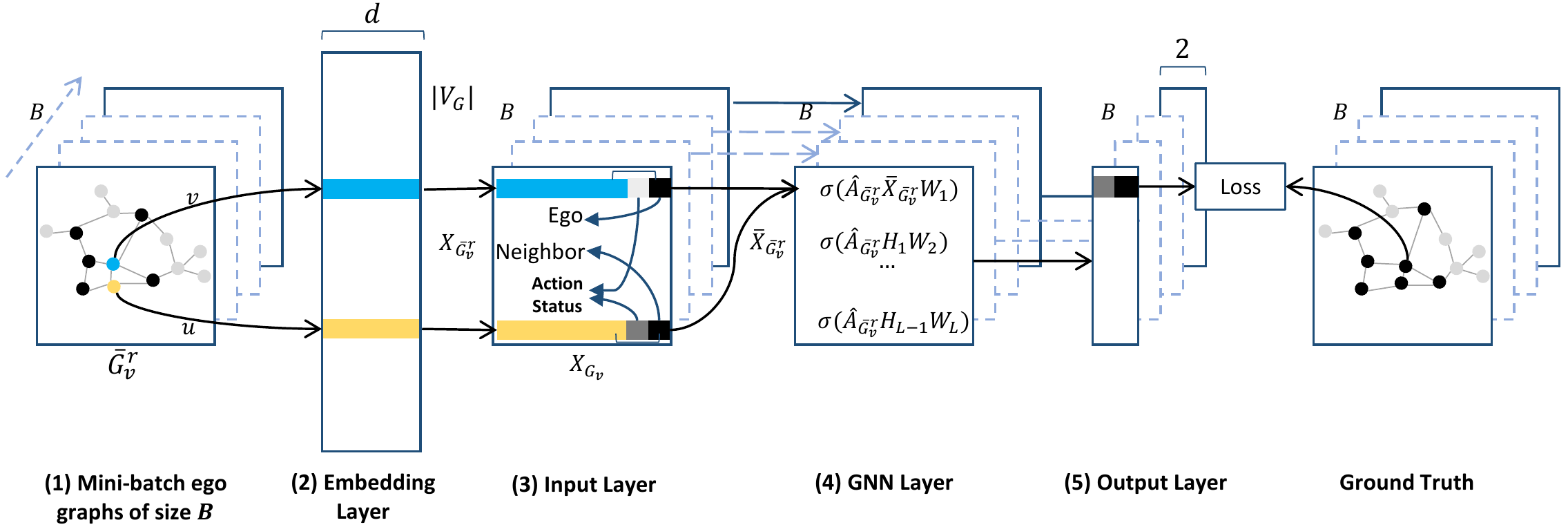}}
\caption{A summary of the DeepInf framework, slightly modified for this work. The node $v$ (blue) is the ego-node while the node $u$ (orange) is a neighbor. For the original architecture, please refer to \citep{Qiu2018}.}
\label{fig:DeepInfarchitecture}
\end{figure*}
\vspace{0.5cm}

\noindent 
Below is a summary that enumerates the steps of the DeepInf framework:

(1) In the first step, DeepInf samples nodes from the graph and extracts their local neighborhoods (i.e., ego-graphs) by using the technique Random walk with restart (RWR) \citep{Tong2006}. The ego-graphs are then sampled to have an equal number of nodes and balanced Action Statuses. Subsequently, the ego-graphs are fed into the model in random batches, which accelerates the training and introduces stochasticity to the learning.

(2) Next, an \textit{Embedding Layer} generates feature representations for the nodes. In our work, we use the same pre-trained features provided by \citep{Qiu2018}.

(3) In the \textit{Input Layer}, the features of each node are concatenated with two extra fields. One field determines the Action Status of the node, and the other determines the type of the node, i.e., whether this is the central node (ego-node) or a neighbor node.
\newpage
(4) In this step, a \textit{GNN Layer} builds a representation of the nodes using both the structure and embedding features of the ego-graph as input. In \citep{Qiu2018}, two variations of the GNN Layer were proposed, DeepInf-GCN, a GNN that employs the spectral rule \citep{Kipf2017} and DeepInf-GAT, a GNN that uses multi-head attention when aggregating the features of the node \citep{Velickovic2018}. In this work, we focus on improving the performance of DeepInf-GCN

(5) Finally, the \textit{Output Layer} predicts the action status of the central node, minimizing the following loss function:

\begin{equation}
  \mathcal{L}(\Theta) = -\sum_{i=1}^{N}log(P_{\Theta}(s_{v_i}^{t+\Delta t}|\bar{G}_{v_i}^r,\bar{S}_{v_i}^t))
\end{equation}

Therefore, the problem of predicting the user-level influence is reduced to a binary classification problem that is solved by minimizing the negative log-likelihood for the Action Status of a node at time $t+\Delta t$ w.r.t to the model parameters $\Theta$. Where $\bar{G}_{v_i}^r$ is the graph structure of the sampled neighborhood of node $v_i$ at time $t$, and $\bar{S}_{v_i}^t$ is the Action Status of the neighborhood of node $v_i$ at time $t$.

Figure \ref{fig:DeepInfarchitecture} represents a summary of the DeepInf framework. Note that in our work, two steps were removed from the original framework: The first is \textit{Instance Normalization}. This step normalizes the embeddings with learnable parameters to avoid overfitting, as described in \citep{Ulyanov2016}. The effect of such parametrized normalization on the preservation of the topological features is beyond the scope of this study, therefore this step was omitted. The second is concatenating \textit{Hand-crafted Centrality Features} to the pre-trained embeddings. We forgo this step since we want to test the ability of the models to predict the influence, by leveraging their own capacity at automatically encoding the centrality features, rather than using manually defined ones. More details about the original framework can be found in \citep{Qiu2018}.

\subsection{Datasets and Compared Baselines}

Motivated by the fact that the social influence of a node is highly connected to some of its topological features, such as degree, eigenvector, and betweenness centrality \citep{Li2016}, we leverage the finding of our work — that GAE\_CONCAT is capable of capturing these important centrality measures in its embedding — and propose DeepInf-CONCAT, a simple variation to the DeepInf-GCN model. As we will show, our experimental results illustrate that DeepInf-CONCAT significantly improves the prediction capacity of the original model, without the need for explicitly introducing hand-crafted features in the training. In the following, we describe DeepInf variations that we have considered in our comparative analysis.

\begin{itemize}
\item  \textbf{DeepInf-CONCAT-64*} is a variation of with two layers that uses the SUM aggregation rule and concatenates the output of the first and second layer as embedding, equivalent to GAE\_CONCAT. 

\item  \textbf{DeepInf-L1-SUM-128} is a variation with one layer that uses the SUM aggregation rule, equivalent to GAE\_L1\_SUM.  

\item  \textbf{DeepInf-L2-SUM-128} is a variation with two layers that uses the SUM aggregation rule and the output of the second layer as embedding, equivalent to GAE\_L2\_SUM. 

\item  \textbf{DeepInf-MEAN-128} is a variation with two layers that uses the MEAN aggregation rule with the output of the second layer as embedding, equivalent to GAE\_MEAN.

\item  \textbf{DeepInf-GCN-128} is the same model presented in \citep{Qiu2018}. It aggregates the messages using the spectral rule and uses the output of the last layer as embedding, equivalent to GAE\_SPECTRAL.
\end{itemize}
\newpage

For the hidden layers of the models, we use the hyperbolic tangent activation function for the new variations, and a hidden layer of size 128, except for DeepInf-CONCAT-64 where we use a hidden layer of size  64. It is important to note that given the nature of the DeepInf model as a GNN rather than GAE, implementing an equivalent to GAE\_FIRST or GAE\_MIXED is not possible. Furthermore, since the effect of multi-head attention on the preservation of topological features is beyond the scope of this work, and for a fairer and clearer comparison between the other models that do not employ attention, we do not include the results of DeepInf-GAT.

\begin{table}[!t]
\centering
\fontsize{8}{8}\selectfont
\renewcommand{\arraystretch}{1.2}
\centering
\begin{tabular}{lccc}
\hline\hline
\multicolumn{1}{c}{\textbf{Dataset}} & \textbf{Nodes} & \textbf{Edges} & \textbf{Observations} \\ \hline
Digg & 279,630 & 1,548,126 & 24,428 \\
Twitter & 456,626 & 12,508,413 & 499,160 \\
OAG & 953,675 & 4,151,463 & 499,848 \\
Weibo & 1,776,950 & 308,489,739 & 779,164 \\ \hline\hline
\end{tabular}
\caption{Graph Datasets for the DeepInf experiments - The field \textit{Observations} indicates the number of sampled ego-graphs. The figures are retrieved from \citep{Qiu2018}}.
\label{tab:DeepInfdatasets}
\end{table}

To properly measure the effect of the proposed variations on the DeepInf-GCN model, we adopt the same benchmarks in \citep{Qiu2018}: Digg, Twitter, OAG, and Weibo, with the same experimental and training setup. Table \ref{tab:DeepInfdatasets} summarizes the details of the four datasets.

\subsection{Prediction Results and Discussion}
\begin{table}[!t]
\centering
\fontsize{8}{8}\selectfont
\renewcommand{\arraystretch}{1.2}
\centering
\begin{tabular}{clcccc}
\hline\hline
\textbf{Dataset} & \multicolumn{1}{c}{\textbf{Model}} & \textbf{AUC} & \textbf{PREC} & \textbf{REC} & \textbf{F1} \\ \hline
\multicolumn{1}{c|}{\multirow{6}{*}{OAG}} & DeepInf-CONCAT-64 & \textbf{0.675} & 0.338 & 0.672 & \textbf{0.450} \\
\multicolumn{1}{c|}{} & DeepInf-L1-SUM-128 & 0.651 & 0.327 & 0.660 & 0.438 \\
\multicolumn{1}{c|}{} & DeepInf-L2-SUM-128 & 0.671 & \textbf{0.348} & 0.632 & 0.449 \\
\multicolumn{1}{c|}{} & DeepInf-MEAN-128 & 0.652 & 0.331 & 0.647 & 0.438 \\
\multicolumn{1}{c|}{} & DeepInf-GCN-128 & 0.613 & 0.289 & \textbf{0.746} & 0.417 \\
\multicolumn{1}{c|}{} & DeepInf-GCN-128-HF & 0.635 & 0. 302 & 0.743 & 0.430 \\ \hline
\multicolumn{1}{c|}{\multirow{6}{*}{Digg}} & DeepInf-CONCAT-64 & \textbf{0.881} & \textbf{0.722} & 0.718 & \textbf{0.720} \\
\multicolumn{1}{c|}{} & DeepInf-L1-SUM-128 & 0.874 & 0.686 & 0.716 & 0.701 \\
\multicolumn{1}{c|}{} & DeepInf-L2-SUM-128 & 0.874 & 0.718 & 0.692 & 0.705 \\
\multicolumn{1}{c|}{} & DeepInf-MEAN-128 & 0.804 & 0.606 & 0.675 & 0.639 \\
\multicolumn{1}{c|}{} & DeepInf-GCN-128 & 0.866 & 0.680 & \textbf{0.720} & 0.699 \\
\multicolumn{1}{c|}{} & DeepInf-GCN-128-HF & 0.841 & 0.587 & 0.676 & 0.628 \\ \hline
\multicolumn{1}{c|}{\multirow{6}{*}{Twitter}} & DeepInf-CONCAT-64 & \textbf{0.783} & \textbf{0.473} & 0.665 & \textbf{0.553} \\
\multicolumn{1}{c|}{} & DeepInf-L1-SUM-128 & 0.762 & 0.442 & 0.671 & 0.533 \\
\multicolumn{1}{c|}{} & DeepInf-L2-SUM-128 & 0.772 & 0.448 & \textbf{0.675} & 0.539 \\
\multicolumn{1}{c|}{} & DeepInf-MEAN-128 & 0.727 & 0.408 & 0.624 & 0.494 \\
\multicolumn{1}{c|}{} & DeepInf-GCN-128 & 0.768 & 0.459 & 0.633 & 0.532 \\
\multicolumn{1}{c|}{} & DeepInf-GCN-128-HF & 0.766 & 0.443 & 0.667 & 0.532 \\ \hline
\multicolumn{1}{c|}{\multirow{6}{*}{Weibo}} & DeepInf-CONCAT-64 & \textbf{0.810} & \textbf{0.472} & 0.732 & \textbf{0.574} \\
\multicolumn{1}{c|}{} & DeepInf-L1-SUM-128 & 0.781 & 0.445 & 0.708 & 0.547 \\
\multicolumn{1}{c|}{} & DeepInf-L2-SUM-128 & 0.796 & 0.444 & \textbf{0.742} & 0.556 \\
\multicolumn{1}{c|}{} & DeepInf-MEAN-128 & 0.774 & 0.433 & 0.723 & 0.542 \\
\multicolumn{1}{c|}{} & DeepInf-GCN-128 & 0.753 & 0.410 & 0.707 & 0.519 \\
\multicolumn{1}{c|}{} & DeepInf-GCN-128-HF & 0.768 & 0.424 & 0.713 & 0.532 \\ \hline\hline
\end{tabular}
\caption{Prediction results on the four datasets for the five variations of DeepInf-GCN.}
\label{tab:DeepInfpredictionperformance}
\end{table}
\begin{table}[!t]
\centering
\fontsize{9}{9}\selectfont
\renewcommand{\arraystretch}{1.2}
\centering
\begin{tabular}{lcccc}
\hline\hline
\multicolumn{1}{c}{\textbf{Model}} & \textbf{OAG} & \textbf{Digg} & \textbf{Twitter} & \textbf{Weibo} \\ \hline
DeepInf-CONCAT-64 & 0.675 & 0.881 & 0.783 & 0.810 \\
DeepInf-GCN-128 & 0.613 & 0.866 & 0.768 & 0.753 \\ \hline
\textbf{Relative Gain} & \textbf{10.1\%} & \textbf{1.7\%} & \textbf{2.0\%} & \textbf{7.6\%} \\ \hline\hline
DeepInf-GCN-128-HF & 0.635 & 0.841 & 0.766 & 0.768 \\ \hline
\textbf{Relative Gain} & \textbf{6.3\%} & \textbf{4.8\%} & \textbf{2.2\%} & \textbf{5.5\%} \\ \hline\hline
\end{tabular}
\caption{Relative gain of DeepInf-CONCAT-64 in terms of AUC against DeepInf-GCN-128 and DeepInf-GCN-128-HF.}
\label{tab:DeepInfRelativeGain}
\end{table}

We report four metrics for evaluating the performance of the models: Area Under Curve (AUC), Precision (REC), Recall (REC), and F1-Measure (F1). Table \ref{tab:DeepInfpredictionperformance} lists the performance of the compared models on the four datasets. Note that we have also included the results of DeepInf-GCN as reported in the original paper (DeepInf-GCN-128-HF). That is DeepInf-GCN trained with the inclusion of the hand-crafted features and the instance normalization trick. The results clearly show that DeepInf-CONCAT-64 outperforms the other models for both AUC and F1 on all four benchmarks. Furthermore, as shown in Table \ref{tab:DeepInfRelativeGain}, DeepInf-CONCAT-64 significantly outperforms DeepInf-GCN-128 with 10.1\% and 7.6\% gains on both OAG and Weibo datasets respectively.

This boost in performance can be attributed in part to the ability of DeepInf-CONCAT to automatically encode topological features that are relevant to the node-level influence in the graph. This observation can be further backed by the fact that the other variations, which also use the SUM aggregation rule, underperformed DeepInf-CONCAT. In other words, the gains of DeepInf-CONCAT are partly due to the concatenation of the first and second layers to form the embedding and the relevant topological information that they encode. As shown in table \ref{tab:DeepInfRelativeGain}, DeepInf-CONCAT-64 also outperforms DeepInf-GCN-128-HF on all four benchmarks, highlighting the importance of allowing an automatic encoding of the relevant topological features vs manually defining them.

\begin{table}[!t]
\centering
\fontsize{8}{8}\selectfont
\renewcommand{\arraystretch}{1.2}
\centering
\begin{tabular}{clcccc}
\hline\hline
\textbf{Dataset} & \multicolumn{1}{c}{\textbf{Model}} & \textbf{AUC} & \textbf{PREC} & \textbf{REC} & \textbf{F1} \\ \hline
\multicolumn{1}{c|}{\multirow{2}{*}{OAG}} & DeepInf-CONCAT-64 & \textbf{0.675} & 0.338 & \textbf{0.672} & 0.450 \\
\multicolumn{1}{c|}{} & DeepInf-CONCAT-128 & 0.674 & \textbf{0.342} & 0.663 & \textbf{0.451} \\ \hline
\multicolumn{1}{c|}{\multirow{2}{*}{Digg}} & DeepInf-CONCAT-64 & 0.881 & 0.722 & 0.718 & 0.720 \\
\multicolumn{1}{c|}{} & DeepInf-CONCAT-128 & \textbf{0.885} & \textbf{0.738} & \textbf{0.723} & \textbf{0.730} \\ \hline
\multicolumn{1}{c|}{\multirow{2}{*}{Twitter}} & DeepInf-CONCAT-64 & 0.783 & \textbf{0.473} & \textbf{0.665} & \textbf{0.553} \\
\multicolumn{1}{c|}{} & DeepInf-CONCAT-128 & \textbf{0.785} & 0.470 & 0.659 & 0.549 \\ \hline
\multicolumn{1}{c|}{\multirow{2}{*}{Weibo}} & DeepInf-CONCAT-64 & 0.810 & 0.472 & \textbf{0.732} & 0.574 \\
\multicolumn{1}{c|}{} & DeepInf-CONCAT-128 & \textbf{0.811} & \textbf{0.474} & 0.728 & \textbf{0.575} \\ \hline\hline
\end{tabular}
\caption{Comparison between the results of DeepInf-CONCAT with hidden layers of size 64 and hidden layers of size 128.}
\label{tab:DeepInfCompare64by128}
\end{table}

It is important to note that DeepInf-CONCAT-64 outperforms the other models while having half their hidden layer size. To study the effect of the hidden layer size and eliminate \textit{overparameterization} as a potential cause for the underperformance of the other models, we compare DeepInf-CONCAT-64 to DeepInf-CONCAT-128, which constitutes the same model with a hidden layer size of 128. As shown in table \ref{tab:DeepInfCompare64by128}, with DeepInf-CONCAT-128, we notice an even higher boost in performance on most of the datasets, further demonstrating the effectiveness of the DeepInf-CONCAT model.

To summarize, in this section, we have demonstrated how we can put to practice the findings of our work concerning the representational power of graph embeddings. We have shown that by carefully choosing an architecture that has been experimentally shown to preserve task-relevant topological features, we are able to significantly improve the model's performance. We believe that such an understanding of the representational power of the GNN and the specialization of each layer in preserving the topological features can greatly reduce the time of the model-search step in the development process and architecture design.

\section{Conclusion}
In this paper, we investigated the representational power of the graph embeddings at preserving important topological structures in the graph. For this purpose, we conducted an extensive empirical study on three unsupervised classes of embedding models, with a special emphasis on Graph Autoencoders. Our results show that five topological features: The Degree, the Local Clustering Score, the Betweenness Centrality, the Eigenvector Centrality, and Triangle Count are concretely preserved in the first layer of the Graph Autoencoder that employs the SUM aggregation rule, under the condition that the model preserves the second-order proximity. We further support our claims with 2D visualizations that reveal a well-organized hierarchical distribution of the node degrees in the embeddings of the aforementioned model. Furthermore, we studied the effect of topological features preservation on downstream tasks. Our results show that the presence of the topological features in the embeddings can significantly enhance the performance of the model on downstream tasks, especially when the preserved topological features are relevant to the target task. Finally, we put our findings to practice in a case study that involved applying our recommendations to a social influence prediction framework. Our results show a great boost in performance, highlighting the importance of having a “Topological Features Preservation” strategy when designing a GNN architecture. In our continuing research, we intend to expand the scope of our empirical study by analyzing other graph structures, such as investigating how the temporal aspects of the dynamic graphs are encoded in the embeddings. This work is underway.

\section*{Acknowledgements}
The  authors  gratefully  thank  the  reviewers  and  the  associate editor for their valuable comments and important suggestions. This work is supported by research grants from the Natural Sciences and Engineering Research Council of Canada (NSERC).

\bibliographystyle{elsarticle-harv} 
\bibliography{References.bib}

\end{document}